\documentclass[preprint,12pt]{elsarticle}

\usepackage{amssymb}
\usepackage[version=4]{mhchem}

\usepackage[utf8]{inputenc}
\usepackage[T1]{fontenc}
\usepackage{subcaption}
\usepackage{nomencl}
\makenomenclature
\usepackage{breqn}
\usepackage{etoolbox}
\usepackage{multirow}
\usepackage{amsmath}

\usepackage{times}  
\usepackage{helvet} 
\usepackage{courier}  
\usepackage[hyphens]{url}  
\usepackage{graphicx} 
\usepackage{graphicx}  
\usepackage{makecell}
\usepackage{comment}
\usepackage[title]{appendix}
\usepackage{hyperref}
\usepackage{siunitx}
\usepackage{nomencl}
\usepackage{xcolor}

\makenomenclature

\makeatletter
\def\thenomenclature{%
  \@ifundefined{chapter}%
  {
    \section{\nomname}
    \if@intoc\addcontentsline{toc}{section}{\nomname}\fi%
  }%
  {
    \chapter{\nomname}
    \if@intoc\addcontentsline{toc}{chapter}{\nomname}\fi%
  }%

  \nompreamble
  \list{}{%
    \labelwidth\nom@tempdim
    \leftmargin\labelwidth
    \advance\leftmargin\labelsep
    \itemsep\nomitemsep
    \let\makelabel\nomlabel}}
\makeatother

\ExplSyntaxOn

\NewDocumentCommand{\choosenomgroup}{m}
 {
  \str_case:nn { #1 }
   {
    {O}{OBD~Parameters}
    {P}{Physics~Model~Variables}
    {A}{Acronyms}
   }
 }
 
\ExplSyntaxOff
\renewcommand*{\nompreamble}{\markboth{\nomname}{\nomname}}

\newcommand{\nomunit}[1]{%
  \renewcommand{\nomentryend}{\hspace*{\fill}#1}%
}

\makeatletter
\def\ps@pprintTitle{%
  \let\@oddhead\@empty
  \let\@evenhead\@empty
  \def\@oddfoot{\footnotesize Preprint submitted to arXiv\hfill}%
  \let\@evenfoot\@oddfoot}
\makeatother

\begin{document}

\begin{frontmatter}




\title{Physics-based machine learning framework for predicting NOx emissions from compression ignition engines using on-board diagnostics data}




\author{Harish Panneer Selvam\fnref{a}}
\author{Bharat Jayaprakash\fnref{a}}
\author{Yan Li\fnref{b}}
\author{Shashi Shekhar\fnref{b}}
\author{William F. Northrop\fnref{a}\corref{cor1}}

\cortext[cor1]{Corresponding author: wnorthro@umn.edu}
\fntext[a]{Department of Mechanical Engineering, University of Minnesota, Minneapolis, MN 55455}
\fntext[b]{Department of Computer Science and Engineering, University of Minnesota, Minneapolis, MN 55455}

\hypersetup{
    pdfauthor={Harish Panneer Selvam, Bharat Jayaprakash, Yan Li, Shashi Shekhar, William F. Northrop},
    pdftitle={Physics-based machine learning framework for predicting NOx emissions from compression ignition engines using on-board diagnostics data},
    pdfkeywords={data-driven, spatiotemporal, NOx prediction, co-occurrence, emission model, machine learning}
}

\begin{abstract}
This work presents a physics-based machine learning framework to predict and analyze oxides of nitrogen (NO\textsubscript{x}) emissions from compression-ignition engine-powered vehicles using on-board diagnostics (OBD) data as input. Accurate NO\textsubscript{x} prediction from OBD datasets is difficult because NO\textsubscript{x} formation inside an engine combustion chamber is governed by complex processes occurring on timescales much shorter than the data collection rate. Thus, emissions generally cannot be predicted accurately using simple empirically derived physics models. Black box models like genetic algorithms or neural networks can be more accurate, but have poor interpretability. The transparent model presented in this paper has both high accuracy and can explain potential sources of high emissions. The proposed framework consists of two major steps: a physics-based NO\textsubscript{x} prediction model combined with a novel Divergent Window Co-occurrence (DWC) Pattern detection algorithm to analyze operating conditions that are not adequately addressed by the physics-based model. The proposed framework is validated for generalizability with a second vehicle OBD dataset, a sensitivity analysis is performed, and model predictions are compared with that from a deep neural network. The results show that NO\textsubscript{x} emissions predictions using the proposed model has around $55\%$ better root mean square error, and around $60\%$ higher mean absolute error compared to the baseline NO\textsubscript{x} prediction model from previously published work. The DWC Pattern Detection Algorithm identified low engine power conditions to have high statistical significance, indicating an operating regime where the model can be improved. This work shows that the physics-based machine learning framework is a viable method for predicting NO\textsubscript{x} emissions from engines that do not incorporate NO\textsubscript{x} sensing.  
\end{abstract}




\end{frontmatter}




\renewcommand{\nomname}{Abbreviations}
\printnomenclature[1in]

\newpage

\section{Introduction}
\label{intro}
 Air pollution from energy use causes over 100,000 annual deaths in the U.S. alone. Nitrogen oxides (NO\textsubscript{x}) are a primary regulated pollutant from internal combustion engines employed in most vehicles on the road. Forced by ever-stringent governmental rules, combustion and aftertreatment technologies have been employed to steadily decrease NO\textsubscript{x} emissions from engine-powered vehicles. Reduced travel in 2020 due to the COVID-19 pandemic led to a further 20\% decrease in Nitrogen Dioxide (NO\textsubscript{2}) in the atmosphere \citet{streiff_2020}, showing the immense impact that combustion-powered vehicles have on the environment. Increasing electrification in the transportation sector is unlikely to abate the greater use of NO\textsubscript{x}-producing compression ignition (CI) engine-powered vehicles, especially in the commercial vehicle sector. 

 NO\textsubscript{x} primarily refers to two gases: nitric oxide (NO) and nitrogen dioxide (NO\textsubscript{2}). They are formed as a byproduct of combustion in CI engines due to high-temperature chemistry involving oxygen and nitrogen in fuel-lean areas of the diffusion flame. NO is predominantly formed during combustion through high-temperature mechanisms, such as the Zeldovich mechanism \cite{heywood1988internal}. In contrast, NO\textsubscript{2} can form both during combustion through reactions involving radicals like HO\textsubscript{2} and after combustion via NO oxidation. At high temperatures, NO\textsubscript{2} is thermodynamically unstable and tends to decompose back into NO. The formation and interconversion of NO and NO\textsubscript{2} are dynamic processes that occur on millisecond timescales, making their study critical for understanding and controlling NO\textsubscript{x} emissions. Data required to accurately predict NO\textsubscript{x} in production engines operating on the road is difficult to obtain because in-cylinder gas temperature cannot be measured at fast enough timescales, nor can it be measured with high enough spatial resolution to capture diffusion flame structures.  
 
Mitigation measures like exhaust gas recirculation (EGR) and selective catalytic reduction (SCR) are used to lower NO\textsubscript{x} formed during combustion to meet regulatory standards. In many diesel vehicles, a diesel oxidation catalyst (DOC) is used upstream of the SCR not only to reduce CO and HC emissions but also to tune the NO:NO\textsubscript{2} ratio \cite{doc2011}. The DOC oxidizes a portion of the NO to NO\textsubscript{2}, resulting in tailpipe NO\textsubscript{x} emissions that contain a higher fraction of NO\textsubscript{2} than the engine-out emissions \cite{carslaw2019diminishing}. However, obtaining data to accurately predict NO\textsubscript{x} in production engines remains challenging because in-cylinder gas temperatures cannot be measured at sufficiently fast timescales or with high enough spatial resolution to capture the diffusion flame structures. Although electrochemical sensors are used in modern CI engines to measure NO\textsubscript{x} concentrations before and after aftertreatment catalysts, low computational cost predictive models can be valuable to detect sensor faults. If prediction algorithms are sufficiently accurate, NO\textsubscript{x} prediction models could even be used in lieu of of sensors, saving considerable cost and mechanical complexity. 
 
 On-board diagnostics (OBD) data refers to time-series measurements of various engine and vehicle parameters reported by sensors already installed in production vehicles for control and diagnostics purposes. Whereas many previous studies have used external analyzers mounted on the vehicle in a laboratory to develop emissions prediction models, on-board sensors have improved in accuracy and have great potential to be used to accurately predict and analyze NO\textsubscript{x} emissions at low additional cost.
 

This paper describes and validates a framework for accurately predicting instantaneous engine-out NO\textsubscript{x} emissions from sparse time-series OBD\nomenclature[A]{OBD}{On-board diagnostics} datasets collected from CI engine-powered vehicles using physics-guided, transparent, and interpretable artificial intelligence (AI) \nomenclature[A]{AI}{Artificial Intelligence} methods. The paper also aims to demonstrate a method for predicting emissions from vehicle data at higher speed and lower computational cost than higher fidelity emissions simulation models. The developed framework consists of a physics-based model like those presented in previous work and a divergent window co-occurrence pattern detection algorithm to identify vehicle operating conditions that may not be adequately explained by the physics-based model. The work also includes a secondary validation experiment with an additional dataset to prove generalizability, a sensitivity analysis to study effect of the dependent parameters, and a comparison of physics-based model predictions with a black-box artificial neural network (ANN) \nomenclature[a]{ANN}{Artificial neural network} model.

\section{Background}
Notable works in the literature related to modeling and predicting vehicular NO\textsubscript{x} emissions can be categorized based on whether they use data from engine or vehicle experiments, conducted in a laboratory or from sensors mounted on in-use vehicles. Prediction models can be further classified into empirical and black-box approaches. Empirical models use physical relations to predict emissions values, while black-box models use purely numerical or statistical means to predict NO\textsubscript{x} from other collected variables. Figure 1 illustrates the taxonomy of NO\textsubscript{x} prediction models in the literature and illustrates where the current work fits within this landscape.

\begin{figure}[h!]
    \centering
    \includegraphics[width=0.5\linewidth]{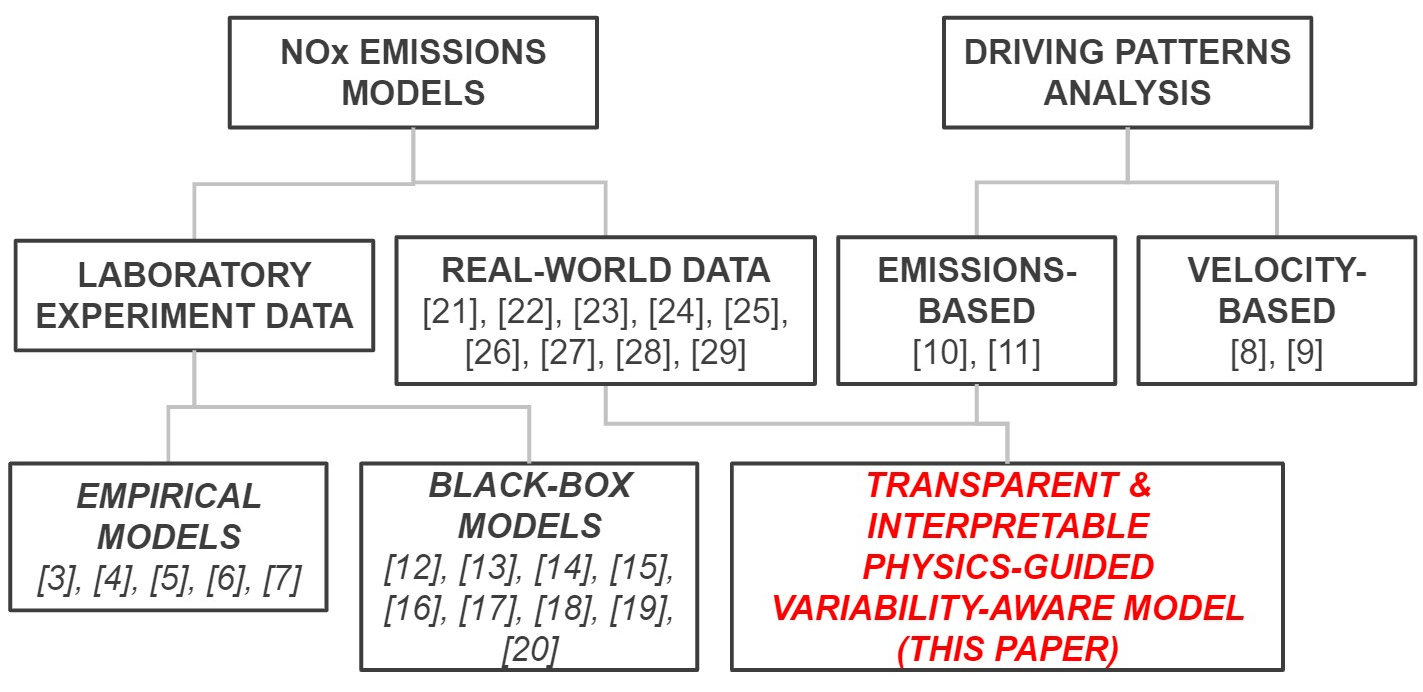}
    \caption{Proposed taxonomy to compare related works in literature with this paper}
    \label{fig:taxonomy}
\end{figure}

Laboratory data obtained primarily from engine or chassis dynamometer experiments are useful for creating well-defined empirical or phenomenological models that predict emissions at specific engine operating conditions. For example, \citet{Gao2001}, and \citet{Molina2014} proposed phenomenological models that describe NO\textsubscript{x} formation from engines under fixed laboratory conditions. \citet{Ozgul2019} developed a fast NO\textsubscript{x} emission prediction methodology by utilizing one-dimensional models generated in commercial engine simulation software using the Extended Zeldovich mechanism included in the package. The model was then tuned by embedding different calibration multiplier maps. The simplicity of such models make them useful in control-related and optimization-based applications (\citet{Asprion2013a}, \citet{Vihar2018}, \citet{Kesgin2003}). Park et al. \cite{park2013prediction} proposed a simple empirical framework to predict NO concentrations in real time using in-cylinder pressure and other data available to the engine control unit. However, the complexity and cost of equipping production engines with in-cylinder pressure sensors present a drawback to adoption.  Although these models effectively describe emissions under controlled operating conditions, they may not be extendable to large, sparse OBD datasets collected from in-use vehicles. 

To increase accuracy and lower computational cost, machine learning models have been proposed for predicting NO\textsubscript{x} from engines. For example, black-box approaches such as ANN models are popular tools employed in recent work related to emissions and energy prediction. The term 'black-box' describes AI models that are not transparent and are difficult to be interpreted by humans. ANN models have been used to predict vehicle emissions like NO\textsubscript{x}, soot, and other pollutants using laboratory data from CI engines as input (\citet{Zhou1999},\citet{Traver1999},\citet{DeLucas2001}, \citet{Taghavifar2014}, \citet{Taghavifar2016}). \citet{lee_machine_2021} conducted real driving emissions (RDE) tests using vehicles equipped with portable emisisons measurement systems to collect data, which was then used to evaluate an black-box ANN model that effectively predicted NO\textsubscript{x} emissions. \citet{li_prediction_2021} proposed a deep-learning differentiation model that used a Singular Spectrum Analysis (SSA) to reduce the noise in the time series data and ICEEMDAN (Improved Complete Ensemble Empirical Mode Decomposition with Adaptive Noise) to decompose the noise-reducing data into several stable subsequences before using GRU (Gated Recurrent Unit) and SVR (Support Vector Regression) for the prediction of the sequences. \citet{wang_remote_2021} used a Long Short-Term Memory neural network to establish a NO\textsubscript{x} prediction model using several components from the OBD data that were selected as inputs to the model using partial least squares (PLS) method. Other related works have used different neural network architectures like back propagation neural networks with mutual information (\citet{Wang2020}) and genetic algorithms (\citet{Alonso2007}) to predict NO\textsubscript{x} emissions. In the literature, ANN models are primarily used as an easy and efficient method to predict NO\textsubscript{x} (\citet{Roy2014}, \citet{Mohammadhassani2012}, \citet{fang_artificial_2021}) given a set of feature variables. A key limitation of previous ANN methods is that they do not include physical relations in the neural networks. Due to their non-transparent black-box nature, it becomes impossible for an engineer to interpret relationships between the results and different feature variables used as inputs. Without a physics basis, the ANN models run the risk of overfitting, especially when the number of data entries is low compared to the number of levels used in the model. Hence, ANN models provide little physical interpretability when attempting to explain underlying reasons for trends in NO\textsubscript{x} emissions as a function of feature variables.


The framework presented in this paper can be classified as a physics-based emissions model with additional functionality developed to identify operating conditions that lead to high predictive errors. One example of a general physics-based model from the literature to predict NO\textsubscript{x} and soot emissions of a diesel engine under warm-up conditions in a laboratory setting is provided by \citet{Tauzia2018}. \citet{Aliramezani2019} proposes a polynomial equation to predict NO\textsubscript{x} and a simple model for finding BMEP based on experimental data for the purpose of a multi-input multi-output(MIMO) engine control. \citet{Fomunung1999} uses ordinary least-square regression like tree-regression to predict load-dependant NO\textsubscript{x} emission values. \citet{Yang2016} uses OBD data from diesel and hybrid buses to predict vehicle-power-dependant NO\textsubscript{x} emissions. \citet{Bostic2009} models NO\textsubscript{x} emissions of an off-road diesel vehicle using a Non-linear Polynomial Network Modeling. \citet{Mata2016} predicts NO\textsubscript{x} and fuel consumption using bi-harmonic maps on data obtained in four specific dynamic bus operating conditions. \citet{Sharma2019} proposed a semi-empirical model for NO\textsubscript{x} emissions on synthetic data obtained from a commercially available engine simulation code package. \citet{Tauzia2017} developed semi-physical models from zero-dimensional thermodynamic models to predict soot and NO\textsubscript{x} emissions and calibrated on standard engine map. The closest work in literature to the research presented here is given by \citet{Rosero2020}, who used in-use data obtained from a portable emission measurement system (PEMS) to predict emissions and fuel efficiency with the help of engine map development method. However, this approach fails to include feature pattern analysis, a step essential for understanding trends in the various operating conditions represented in the dataset. In the authors' previous work, \citet{PanneerSelvam2020}, provides an analytical partition-based approach and co-occurrence pattern detection to obtain more refined NO\textsubscript{x} predictions using a physics-based regression model from an on-board diagnsotics dataset. The work presented here moves away from partitioning, and proposes a way to refine and understand a single model expression for accurately predicting NO\textsubscript{x} from a given engine based on vehicle OBD data with standard feature variables.

\section{Methodology}
\subsection{Proposed Physics-Based Model for \texorpdfstring{NO\textsubscript{x}}~ Emission Prediction}
To address the limitations of both fully empirical and fully black-box approaches present in literature, a physics-based non-linear regression model for accurately predicting NO\textsubscript{x} emissions from on-board diagnostics data is proposed. Main parameters were chosen that affect NO\textsubscript{x} formation in a compression-ignition engine based on the general chemical kinetics equation for NO\textsubscript{x} formation and studying the different attributes present in the OBD dataset.

\subsubsection{Chemical Kinetics of \texorpdfstring{NO\textsubscript{x}}~ Formation}
NO\textsubscript{x} emissions are formed inside a CI engine cylinder under very specific conditions. As previously stated, NO\textsubscript{x} is a combination of NO and NO\textsubscript{2}, the former species primarily a precursor to the second. In general, NO formation requires high temperature (around 1800K), high pressure, and a slightly lean mixture ($\phi \sim 0.9$) of fuel and air. NO primarily occurs during the diffusion period of diesel combustion. The most widely accepted mechanism for NO formation is the well-known Extended Zel'dovich Mechanism (\citet{liviu2010simplified}) as given in Equation \ref{eq:chemistryeq}: 

\begin{dmath}
{\ce{ O + N_{2} <-> NO + N }}\\
{\hspace{-8pt}\ce {N + O_{2} <-> NO + O}}\\
{\hspace{-9pt}\ce {N + OH <->NO + H}}
\label{eq:chemistryeq}
\end{dmath}

While the Extended Zel'dovich Mechanism generally covers NO production, NO\textsubscript{2} is sometimes grouped into the overall kinetic expression due to the consistent nature by which it is formed from NO. The general expression for NO\textsubscript{x} rate of formation is given in Equation \ref{eq:kinetic} (\citet{liviu2010simplified}).
\begin{equation}
\frac{d[NO_x]}{dt}=6*10^6*T_{bz}^{0.5}*exp(\frac{-69090}{T_{bz}})*[O_2]_e^{0.5}*[N_2]_e^{0.5}
\label{eq:kinetic}
\end{equation}
where, $\frac{d[NO_x]}{dt}$  \nomenclature[p]{$\frac{d[NO_x]}{dt}$}{Rate of formation of NO\textsubscript{x} \nomunit{\si{mol/cm\textsuperscript{3}\per\second}}}denotes the rate of formation of NO\textsubscript{x} in mol/cm\textsuperscript{3}/s, $[O_2]_e$ and $[N_2]_e$ \nomenclature[p]{$[O_2]_e$, $[N_2]_e$}{Equilibrium concentration of $O_2$, $N_2$ \nomunit{\si{mol\per cm^3}}}represent the equilibrium concentrations of $O_2$ and $N_2$ in mol/cm3 respectively, and $T_bz$ \nomenclature[p]{$T_bz$}{In-cylinder burn-zone mean temperature \nomunit{\si{\kelvin}}} represents the in-cylinder burn-zone mean temperature in K. 

Although this rate equation provides a theoretical interpretation of the general formation of NO\textsubscript{x} products, it does not take into account the many complex combustion processes that take place inside a compression-ignition engine that contribute to NO\textsubscript{x} production. This is especially true when using OBD datasets as input to the model, as many of the key feature variables in Eq. 2 are not measured or recorded. Temperature, pressure and chemical conditions at the site of NO\textsubscript{x} formation in the combustion chamber drives a need for constitutive relations as a function of known variables to yield an accurate prediction.

\subsubsection{On-Board Diagnostics Data}
\begin{figure}[t!]
    \centering
    \includegraphics[width=0.5\linewidth]{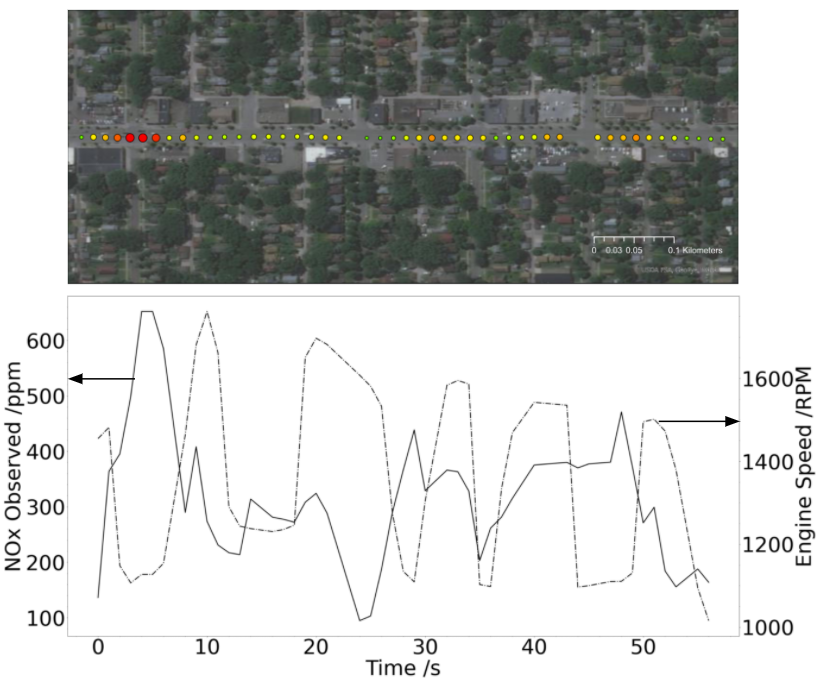}
    \caption{Spatial visualization of a road segment from the transit bus OBD dataset. The colors and sizes of the points in the max correspond to the NO\textsubscript{x} observed value.}
    \label{fig:obd}
\end{figure}

The OBD data used in this study was obtained from a transit bus powered by a Cummins ISB6.7 Diesel hybrid engine system. The bus traversed three different routes over 16 runs in the Minneapolis-St.Paul region during a span of 5 days. The dataset contains 99,895 data entries including measurements of 90 engine and vehicle attributes logged at 1 Hz resolution. The physically-relevant attributes from the OBD system transmitted over the J1939 closed area network on the vehicle and chosen for the proposed framework are given in Table \ref{tab:OBDattributes}. Figure \ref{fig:obd} shows NO\textsubscript{x} and engine rotational speed from the OBD data for the vehicle operating in Minneapolis, MN. The top half of the figure shows a one-minute data stream plotted on a map using ArcGIS Pro with the colors and point sizes denoting the corresponding NO\textsubscript{x} values observed.

\begin{table}[h!]
    \centering
        \caption{List of attributes from the OBD dataset used in the proposed NO\textsubscript{x} emission analysis framework}
    \begin{tabular}{|l|l|l|}
    \hline
    \textit{TimeStamp} & \textit{Date} & \textit{TripIndex}\\
     \textit{engRPM} \nomenclature[o]{\textit{engRPM}}{Engine speed \nomunit{\si{RPM}}} & \textit{EGRkgph} \nomenclature[o]{\textit{EGRkgph}}{Exhaust gas recirculation mass flow rate \nomunit{\si{\kilogram\per\hour}}} & \textit{AirInkgph}\nomenclature[o]{\textit{AirInkgph}}{Intake air mass flow rate \nomunit{\si{\kilogram\per\hour}}} \\
     \textit{intakeT} \nomenclature[o]{\textit{intakeT}}{Intake manifold temperature \nomunit{\si{\kelvin}}}& 
     \textit{intakeP} \nomenclature[o]{\textit{intakeP}}{Intake manifold pressure \nomunit{\si{\kilo\pascal}}}& 
     \textit{Fuelconskgph} \nomenclature[o]{\textit{Fuelconskgph}}{Fuel consumption mass flow rate \nomunit{\si{\kilogram\per\hour}}}\\
     \textit{accelpedalpos} \nomenclature[o]{\textit{accelpedalpos}}{Accelerator pedal position \nomunit{\si{\percent}}}& \textit{ExhaustT}\nomenclature[o]{\textit{ExhaustT}}{Exhaust manifold temperature \nomunit{\si{\kelvin}}} & \textit{WheelSpeed}\nomenclature[o]{\textit{WheelSpeed}}{Vehicle wheel speed \nomunit{\si{\kilo\metre\per\hour}}} \\
     \textit{EngTq}\nomenclature[o]{\textit{EngTq}}{Engine load/torque \nomunit{\si{\newton\metre}}} & \textit{InstFuelEcon}\nomenclature[o]{\textit{InstFuelEcon}}{Instantaneous fuel economy \nomunit{\si{MPG}}} & \textit{SCRingps}\nomenclature[o]{\textit{SCRingps}}{Selective catalytic reduction inlet NO\textsubscript{x} mass flow rate \nomunit{\si{\gram\per\second}}}\\
     \hline
    
    \end{tabular}

    \label{tab:OBDattributes}
\end{table}

\subsubsection{Predictive Accuracy Metrics}
 Accuracy error is likely to be encountered when predicting emissions from a limited number of feature variables using a simplified phenomenological model. In this work, divergence in predicted NO\textsubscript{x} emissions is defined as the difference or deviation of the predicted value from its observed value. Another contributor to divergence is systemic errors, especially with on-board sensors that are designed for low cost and fast time resolution over high accuracy. This drives a need to define divergence from a probabilistic stand point. Here, three predictive accuracy metrics for the purpose are introduced. Coefficient of determination ($R^2$ value)\nomenclature[A]{$R^2$}{Coefficient of determination}, root mean square error (RMSE)\nomenclature[A]{RMSE}{Root mean square error}, and mean absolute error (MAE)\nomenclature[A]{MAE}{Mean absolute error} can be used to effectively evaluate the predictive accuracy of different models. Adjusted $R^2$ value, which is generally used to account for overfitting in a regression problem, is not required owing to the low number of feature variables used in the model. 

\subsubsection{Baseline \texorpdfstring{NO\textsubscript{x}}~ Prediction Model}
The baseline model is taken from \citet{Kumar2018}, who provides an expression for NO\textsubscript{x} emissions (in ppm) using data obtained from an engine dynamometer setup. The assumptions made for this model are that heat transfer is proportional to burnt-gas mass for a constant temperature with negligible changes in specific heat, and that NO\textsubscript{x} formation duration is equal to fuel reaction duration. 

The model obtained by making such assumptions is given by Equation \ref{eq:baseline}:
\begin{equation}
NO_{x,Theoryppm}=A*exp(\frac{-B}{T_{bz}})*\dot{[O_2]}^{0.5}*\frac{t_{comb}}{engRPM}
\label{eq:baseline}
\end{equation}

where $NO_{x,Theoryppm}$ is the predicted NO\textsubscript{x} emissions value in ppm, $\dot{[O_2]}$ \nomenclature[p]{$\dot{[O_2]}$}{Mass flow rate of $O_2$ consumed in combustion \nomunit{\si{\kilogram\per\hour}}}is the mass flow rate of $O_2$ that was consumed during combustion in kg/h,
Tbz is the in-cylinder burnt zone mean temperature (K),
$t_{comb}$ \nomenclature[p]{$t_{comb}$}{Duration of combustion \nomunit{\si{\second}}}
denotes duration of combustion in seconds, $engRPM$ is the engine speed in RPM, and $A$ and $B$ are the calibration factors that depend on engine operating conditions and fuel types used. For a 5.2 L direct injection CI engine, \citet{Kumar2018} estimates the values $A=2.6 *10^7$ and $B = 1100$.

Predictions using the baseline model with constants from \citet{Kumar2018} using the transit bus dataset as input are given in Figure \ref{fig:basepred} and the accuracy metrics for the baseline model are given in Table \ref{tab:accmetrics}. It can be observed that the divergence of NO\textsubscript{x} values with small magnitude are predominantly negative, whereas the larger observed values of NO\textsubscript{x} has highly positive divergence. 


\subsubsection{Regressed Baseline \texorpdfstring{NO\textsubscript{x}}~ Prediction Model}
Model constants A \& B in Equation \ref{eq:baseline} are not appropriate for the specific engine and operating condition reflected in the OBD data. Non-linear regression was used to tune the baseline model for the transit bus dataset. The curve\_fit non-linear regression function from the SciPy Python package  (\citet{2020SciPy-NMeth}) uses Levenberg-Marquardt algorithm to obtain regression coefficients - A \& B values- optimized for the data. Since the dataset consists of numerous vehicle and engine operating conditions, it can be assumed this method would provide probably the best estimate of local minima for the constants. No boundary conditions were enforced for the regression process, and the maximum number of iterations was set to be 500,000. 

\begin{figure*}[htbp!]
     \centering
     \begin{subfigure}[b]{0.45\linewidth}
         \includegraphics[width=\linewidth]{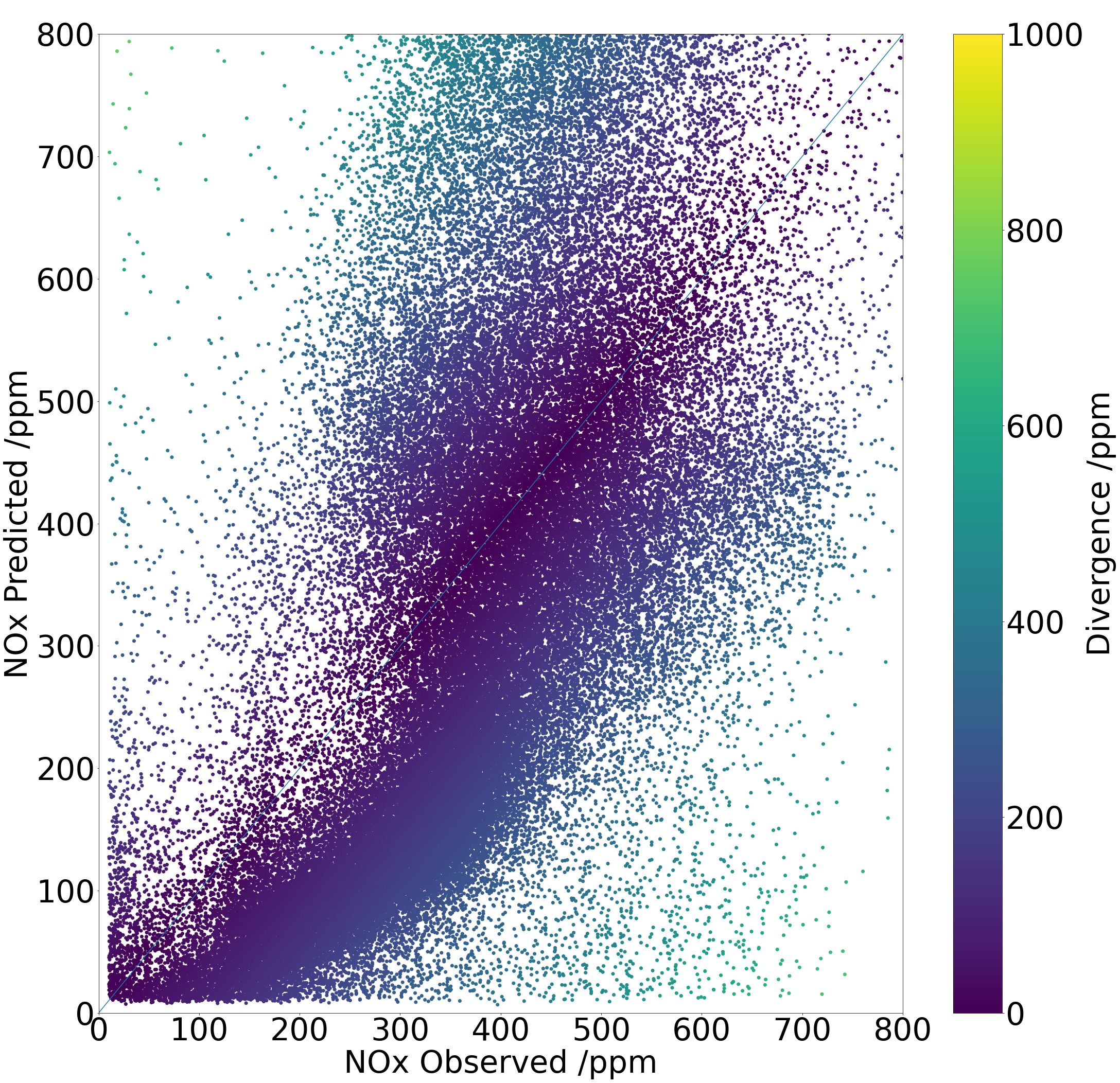}
         \caption{Baseline NO\textsubscript{x} Prediction Model}
         \label{fig:basepred}
     \end{subfigure}
    ~
     \begin{subfigure}[b]{0.45\linewidth}
         \includegraphics[width=\linewidth]{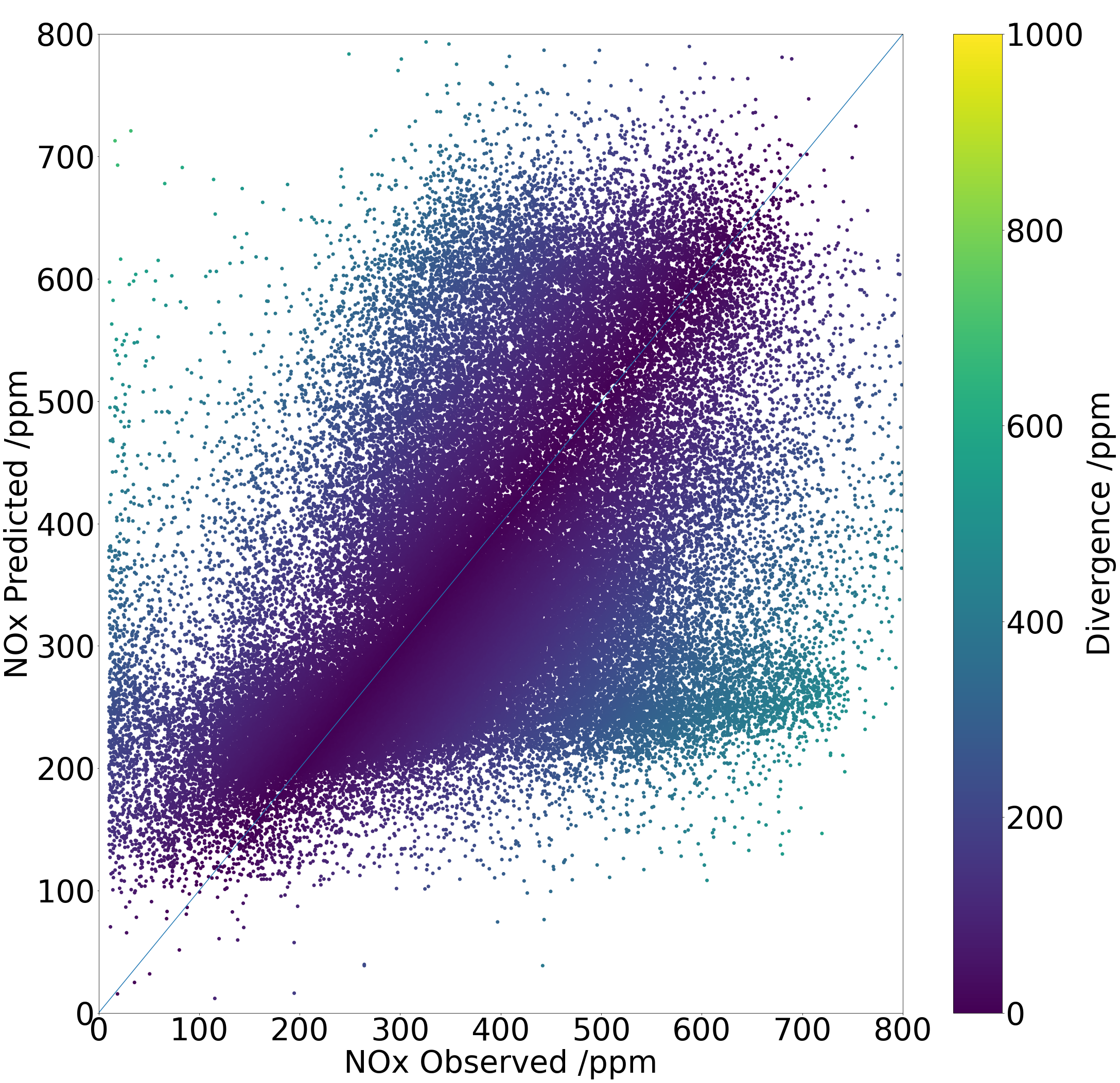}
         \caption{Regressed Baseline NO\textsubscript{x} Prediction Model}
     \end{subfigure}
     \caption{NO\textsubscript{x} Prediction using the Baseline Model and the Regressed Baseline Model}
      \label{fig:baseregpred}
    \end{figure*}

Model constant values obtained from the regressions based on the transit bus dataset were:  $A = 1876153.48$   \& $B = -4168.18$. The prediction using the regressed baseline model is shown in Figure \ref{fig:baseregpred} and the predictive accuracy of the method are given in Table \ref{tab:accmetrics}. When compared to baseline prediction with no modifications shown in \ref{fig:basepred}, the regressed baseline model predictions are more concentrated near the 1:1 line, with divergence values being lower even with the large NO\textsubscript{x} observed values. 


\subsubsection{Proposed \texorpdfstring{NO\textsubscript{x}}~ Prediction Model}
The baseline and the regressed baseline models operate under certain assumptions mentioned in the previous section that contribute to the divergence of points from the 1:1 line. Hence, a new physics-based model was developed to address these assumptions, and potentially provide better predictions when using OBD data as input. The proposed NO\textsubscript{x} prediction model for a compression-ignition engine uses the generalized kinetic equation for the Extended Zel'dovich mechanism given in Equation \ref{eq:kinetic} as its foundation. The Proposed NO\textsubscript{x} prediction model is given in Equation \ref{eq:propphysics}. 

\begin{equation}
    x_{NO_x}=a*\hat{t_{res}}^b*x_{O_2}^c*\hat{T_{adiab}}^d*exp(-e/T_{adiab})
    \label{eq:propphysics}
\end{equation}

where $x_{NO_x}$\nomenclature[p]{$x_{NO_x}, x_{O_2}$}{Mole fraction of NO\textsubscript{x} formed and $O_2$ consumed \nomunit{\si{-}}} is the mole fraction of NO\textsubscript{x} formed,
$\hat{t_{res}}$ \nomenclature[p]{$\hat{t_{res}}$, $\hat{T_{adiab}}$}{Dimensionless form of $t_{res}$, $T_{adiab}$ \nomunit{\si{-}}} is the dimensionless form of residence time which is obtained by multiplying $t_{res}$ \nomenclature[p]{$t_{res}$}{Residence time of combustion \nomunit{\si{\second}}} by engine speed in revolutions per minute (RPM),
$x_{O_2}$\nomenclature[p]{$x_{O_2}$}{Mole fraction of $O_2$ taking part in combustion \nomunit{\si{-}}} is the concentration of oxygen inside the combustion chamber taking part in combustion,
$\hat{T_{adiab}}$ is the dimensionless adiabatic flame temperature of the combustion products($\hat{T_{adiab}}=T_{adiab}/intakeT$), and $a, b, c, d, e$ are the coefficients to be obtained through regression analysis.

The attributes from the OBD data used to predict NO\textsubscript{x} values in the proposed model are engine speed in RPM(\textit{engRPM}), air intake flow rate in kg/h (\textit{AirInkgph}), fuel consumed flow rate in kg/h (\textit{Fuelconskgph}), intake temperature in Kelvin (\textit{intakeT}), intake pressure in Pa (\textit{intakeP}), and EGR flow rate in kg/h (\textit{EGRkgph}). The ground truth values or the target values of $x_{NO_x}$ are obtained from the mass flow rate measured from the inlet to the selective catalytic reduction system in grams per second (\textit{SCRingps}), \textit{AirInkgph} and \textit{Fuelconskgph}. 

{\textbf{Calculating residence time of combustion:}}
The residence time of fuel taking part in diffusion combustion ($t_{res}$) is considered to be equal to the duration of formation of NO\textsubscript{x}. This is because the lean fuel-air mixture used in a compression-ignition engine tries to achieve highly efficient combustion, and hence maintains the high temperature required for NO\textsubscript{x} formation until the end of combustion. Thus, it is calculated $t_{res}$ from the fuel injection duration (\citet{Heywood1979}, \citet{lakshminarayananmodelling}, \citet{TurnsStephenR2000Aitc})



\citet{Alkidas1987} gives an empirical equation to estimate the value of $\beta$ and \citet{10.2307/44658184} provides an equation for ignition delay, which is defined as the time delay between time of injection and the peak pressure during combustion of the fuel. 

\begin{dmath}
\theta_{ID}=(0.36+0.22*MPS)*exp(E_a*(\frac{1}{(R_u*intakeT*CR^{(\gamma-1)})}-\frac{1}{17190})+(\frac{21.2}{(intakeP*CR^{\gamma}-12.4})^{0.63}))\\
{\beta=\frac{0.45*\theta_{ID}*engRPM*60*6}{Fuelconskgph*1000000}}\\
\label{eq:thetaID}
\end{dmath}
\nomenclature[p]{$R_u$ }{Gas constant \nomunit{\si{\joule\per\mole\per\kelvin}}}
\nomenclature[p]{$\beta$ }{Fraction of fuel burned in premixed combustion \nomunit{\si{-}}}
\nomenclature[p]{$E_a$ }{Activation energy \nomunit{\si{\joule}}}
\nomenclature[p]{$\theta_{ID}$ }{Ignition delay \nomunit{\si{\degree}}}
\nomenclature[p]{$MPS$ }{Mean piston speed \nomunit{\si{\metre\per\second}}}
\nomenclature[p]{$CR$ }{Compression ratio \nomunit{\si{-}}}
\nomenclature[p]{$\gamma$}{Polytropic index \nomunit{\si{-}}}

\begin{dmath}
{t_{inj}=\frac{Fuelconskgph*2*1000000}{(6*60*113*835*engRPM)}}\\
{t_{res}=t_{inj}*(1-\beta)}
\label{eq:tres}
\end{dmath}

\nomenclature[p]{$t_{inj}$}{Duration of fuel injection \nomunit{\si{\second}}}
\nomenclature[p]{$\beta$}{Ratio of fuel used in premixed to diffused combustion \nomunit{\si{\second}}}

The fraction of fuel burned in premixed combustion is given by $\beta$, where the residence time of fuel participating in diffusion combustion in seconds is calculated as given in Equation \ref{eq:tres}.

\textbf{Calculating concentration of oxygen taking part in combustion:}
The concentration of oxygen participating in combustion is calculated at an equivalence ratio of $\phi_{NO_x}=0.9$, which is considered as the phi at the combustion plume where NO\textsubscript{x} formation takes place. The stoichiometric chemical equation for diesel combustion is:

\begin{dmath}   
\ce{{C_{13.88}H_{24.06} + 94.74 [0.21{O_{2}} + 0.79{N_{2}}]}}\\
\ce{-> {13.88{CO_{2}} + 12.03{H_2O} + 74.85{N_{2}}}}
\end{dmath}

The mole fraction concentration of oxygen participating in combustion is heavily diluted due to the exhaust gas recirculation (EGR) system, which mixes part of the exhaust gas into the intake manifold in order to reduce the in-cylinder temperature. This diluted charge and the reduced temperature makes it hard for NO\textsubscript{x} to form. The mole fraction of oxygen after charge dilution is calculated from an oxygen balance that considers the intake, engine cylinder, fuel injection and exhaust within a control volume. 

\begin{dmath}
x_{O_2} = \frac{\frac{19.90}{\phi_{NO_x}} + \frac{X_{exh,O_2}*EGRkgph*M_{fuel}}{(0.032*Fuelconskgph)}}{(1+\frac{19.90*4.76}{\phi_{NO_x}}+\frac{EGRkgph}{Fuelconskgph}*(\frac{X_{exh,O_2}}{0.032}+\frac{X_{exh,N_2}}{0.028}+\frac{X_{exh,CO_2}}{0.044}+\frac{X_{exh,H_2O}}{0.018})}
\end{dmath}
where $\phi_{NO_x}$  \nomenclature[p]{$\phi_{NO_x}$ }{Equivalence Ratio of NO\textsubscript{x} \nomunit{\si{-}}} is the equivalence ratio of NO\textsubscript{x} formation $= 0.9$,
$X_{exh,O_2}$, $X_{exh,N_2}$, $X_{exh,CO_2}$, ${X_{exh,H_2O}}$ \nomenclature[p]{$X_{exh,i}$}{Mass fraction of species i=$O_2$, $N_2$, $CO_2$, $H_2O$ in exhaust gas \nomunit{\si{-}}}are the mass fractions of oxygen, nitrogen, $CO_2$ and $H_2O$ found in the exhaust,
$M_fuel$ \nomenclature[p]{$M_{fuel}, M_{product}$}{Molar mass of fuel, product \nomunit{\si{\kilogram\per\mole}}} is the molar mass of the fuel$ = 0.19065 kg/mol$.

\textbf{Calculating adiabatic flame temperature:}
The adiabatic flame temperature is defined as the temperature of the combustion products inside the combustion chamber if all the heat produced during the reactions is used to only increase the temperature of the products. In other words, it is the temperature of the combustion products when there is no heat loss to the cylinder walls and no expansion or compression of gases in the cylinder. The energy balance used to estimate the adiabatic flame temperature is given in Equation \ref{eq:tadiab}:

\begin{dmath}
LHV*M_{fuel}= \sum_{i=CO_2,H_20,N_2,O_2} {n_{i}* \int_{T_{peak}}^{T_{adiab}} C_{p,i} \,dT }
\label{eq:tadiab}
\end{dmath}
where,
LHV \nomenclature[p]{$LHV$}{Lower heating value of diesel fuel \nomunit{\si{\joule\per\kilogram}}} is the lower heating value of diesel fuel is $42.64*10^6$ J/kg,
$n_{i}$ \nomenclature[p]{$n_i$}{Number of moles of $CO_2$, $H_2O$, $N_2$, $O_2$ in products \nomunit{\si{\mole}}}
corresponds to the number of moles of $CO_2$, $H_2O$, $N_2$ and $O_2$ respectively per engine cycle,
$C_{p,i}$ \nomenclature[p]{$C_{p,i}$}{Isobaric specific heats of species i= $CO_2$, $H_20$, $N_2$, $O_2$ \nomunit{\si{\joule\per\kelvin\per\mole}}} are the time dependent specific heat values of $CO_2$, $H_2O$, $N_2$ and $O_2$ respectively in J/K/mol (obtained from look-up tables in \citet{TurnsStephenR2000Aitc}),
$T_{peak}$ \nomenclature[p]{$T_{peak}$}{Peak cylinder temperature \nomunit{\si{\kelvin}}} is the peak temperature of reactants inside the combustion chamber at end of compression stroke in K,
\nomenclature[p]{$P_{peak}$}{Peak cylinder pressure obtained during the top dead center \nomunit{\si{\pascal}}} 
and 
$T_{adiab}$ is the adiabatic flame temperature of the exhaust mixture in K.

{\bf Calculating x\textsubscript{NO\textsubscript{x}}:}
The mole fraction of NO\textsubscript{x} emissions produced in a time step or $x_{NO_x}$ is calculated from SCRingps, Fuelconskgph, and EGRkgph as given in Equation \ref{eq:xNOxActual}

\begin{equation}
x_{NO_{x,Obs}}=\frac{SCRingps*3.6*M_{product}}{0.040*Exhaustkgph}
\label{eq:xNOxActual}
\end{equation}
where $x_{NO_{x,Obs}}$ is the mole fraction of NO\textsubscript{x} emissions observed, SCRingps is the inlet mass flow rate to the selective catalytic reduction in grams per second, Exhaustkgph is the exhaust gas mass flow rate in kilograms per hour, $M_{product}$ is the molar mass of the products (~0.02885 kg/mol).

\begin{figure}[t]
    \centering
    \includegraphics[width=0.5\linewidth]{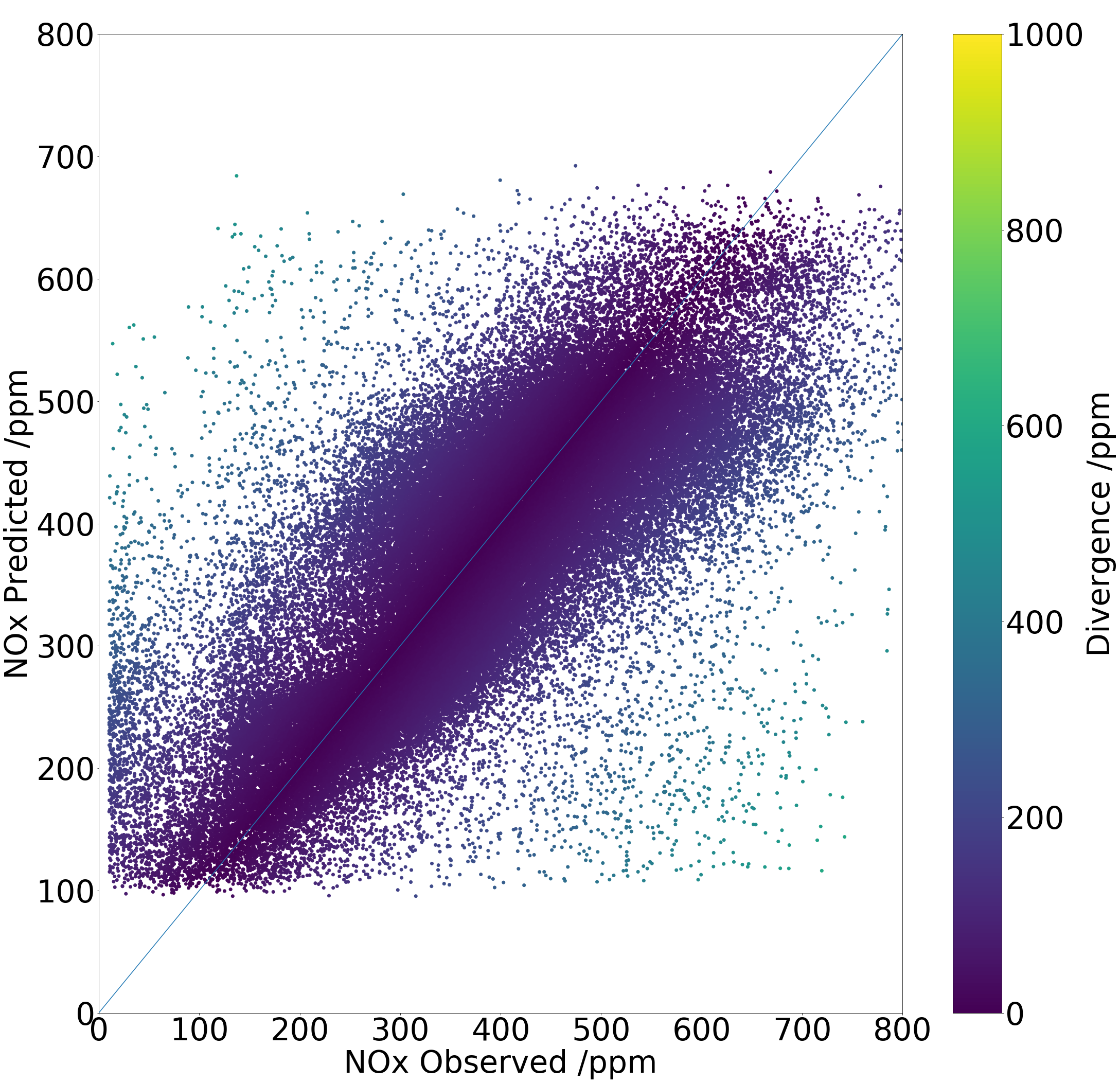}
    \caption{NO\textsubscript{x} Prediction using the proposed physics-based model}
    \label{fig:proppred}
\end{figure}

{\bf NO\textsubscript{x} Value Prediction and Calculation:} Similar non-linear regression is used to obtain all the five regression coefficients and the predicted $x_{NO_x}$ values are converted to ppm according to Equation \ref{eq:NOxTheoryppm} in order to compare with the observed values. The regressed coefficients are
$a=2.15e-3, b=0.013, c=0.36, d=0.082,$ and $e=2117.33 $. The regressed coefficients seem to match the values from literature. For example, the value for '$e$' is similar to the magnitude of 'B' from \ref{eq:baseline}, and values for 'c' and 'd' in Equation \ref{eq:propphysics} are similar to the exponent values for $[O_2]$ and $T$ in Equation \ref{eq:kinetic} respectively. The NO\textsubscript{x} emission values predicted using the proposed physics-basd approach for the transit bus dataset is shown in Figure \ref{fig:proppred} and the accuracy metrics are compared to that of the other models in Table \ref{tab:accmetrics}. From Table \ref{tab:accmetrics} it is illustrated that the proposed physics-based model is more accurate than the baseline and the regressed baseline model in predicting NO\textsubscript{x} from an OBD dataset of a transit bus. 

\begin{equation}
    NO_{x,Theoryppm}=x_{NO_x,Theory}*1000000
    \label{eq:NOxTheoryppm}
\end{equation}
\nomenclature[p]{$NO_{x,Theoryppm}$}{Predicted NO\textsubscript{x} emission value \nomunit{\si{ppm}}}
\nomenclature[p]{$x_{NO_x,Theory}$}{Mole fraction of predicted NO\textsubscript{x} emission value \nomunit{\si{-}}}

\begin{table}[h!]
    \centering
        \caption{Predictive accuracy of Different Models to predict NO\textsubscript{x} emission values}
    \begin{tabular}{l|c|c|c}
    \hline
      \textbf{Model}  & \textbf{R\textsuperscript{2}} & \textbf{RMSE (ppm)} & \textbf{MAE (ppm)}\\
       \hline
       Baseline  & 0.4652 & 204.29 & 177.56\\
       Regressed Baseline & 0.3068 & 133.72 & 97.41\\
       Proposed & 0.6175 & 92.37 & 68.70\\
       \hline
    \end{tabular}

    \label{tab:accmetrics}
\end{table}

\subsection{Divergent Window Co-occurrence Patterns}
It is generally known that an engine operates under a finite number of scenarios which can be defined by similar occurrences of values of measurable engine parameters like engine speed, fuel mass flow rate, and so on. Therefore, a method is proposed here to identify these co-occurrences using a method coined Divergent Window Co-occurrence (DWC)\nomenclature[A]{DWC}{Divergent Window Co-occurrence} Pattern Detection. The method is based on an algorithm developed by \citet{Ali2017}. The primary algorithm takes in OBD data and constraints as an input and outputs all statistically significant DWC patterns.

A divergent window of NO\textsubscript{x} emissions refers to a period \textit{L} seconds in a time series of OBD data records within which the prediction errors (absolute difference between predicted and observed NO\textsubscript{x} value) of the baseline approach exceed an input divergence threshold (\textit{summationThreshold}). A co-occurrence pattern in a time series of OBD data entries is similar to a sequential association pattern except for the use of a spatial statistical interest measure, i.e., Ripley's Cross-K function ($\epsilon$) specialized for time series data. In other words, the pattern represents those subsets of engine attributes and their specific value ranges, which are present together in many divergent time windows and have Cross-K function values above a given threshold ($\epsilon$) and support (i.e., the number of divergent windows in a pattern divided by the total number of time windows) greater than \textit{minsupp}. 

The Cross-K function can be considered as a measure of how well defined a pattern is in a dataset and the support gives an understanding of the frequency of occurrence of a certain pattern. Here, an $\epsilon$ is chosen that is greater than what is obtained if the distribution has complete spatial randomness (CSR). In order words, it is chosen if the pattern is a chance occurrence in the dataset. \textit{minsupp} is defined in a way that the DWC pattern occurs a significant enough number of times to affect the general prediction model, resulting in a prognostic NO\textsubscript{x} emission analysis for the vehicle. 

\subsubsection{Toy Example of a Divergent Window Co-occurrence Pattern}
Table \ref{tab:cooccurrence} shows examples of co-occurrence patterns from the transit busOBD dataset along with their grouping and interpretations. For example, the group "Low Vehicle Speed Condition” has two examples of co-occurrence patterns based on low wheel speed, represented as $w_0$ in a time period of three time points where $w_0$ represents the lowest value range for the $Wheelspeed$ parameter. The magnitude of each attribute is linearly discretized into ten equal intervals represented by the subscripts. The intervals are defined for each attribute between a minimum and maximum values among entries in the dataset. An example is $engRPM$, which is divided into 10 equal intervals between 800 and 1800 RPM for the transit bus dataset. This means, $N_0$ corresponds to entries where $engRPM \epsilon[800,900)$, $N_1$ is for $engRPM \epsilon[900,1000)$, and so on. Similarly, the patterns mined that have very low exhaust gas recirculation (EGR) rate and those during transient events are grouped, since the EGR rate is an important factor influencing $NO_{x}$ formation and transient events cannot be derived from stationary laboratory conditions. 

\begin{table}
   \caption{Examples of Divergent Window Co-occurrence (DWC) patterns and their respective vehicle scenarios. The subscripts for the attributes denote their magnitudes on a linear scale of 0-10. 0,1 corresponds to Very Low values; 2,3,4 to Low values; 5,6,7 to Moderately High values, and 8,9,10 to High values}
    \resizebox{\linewidth}{!}{%
    \centering
     
    \begin{tabular}{m{6cm}|l l}
  	\hline
      \textbf{Scenario} & \textbf{Example Patterns } \\
      \hline
      Low Vehicle Speed Condition &   \makecell{1. \textit{Wheelspeed}:  $w_{0}$ $w_{0}$ $w_{0}$ \\ \textit{IntakeT}: $T_{9}$ $T_{9}$ $T_{9}$} & \makecell{2. \textit{Wheelspeed}:  $w_{0}$ $w_{0}$ $w_{0}$ \\ \textit{IntakeT}: $T_{9}$ $T_{9}$ $T_{9}$\\
      \textit{Fuelconskgph}: $F_{2}$ $F_{2}$ $F_{2}$}\\
      \hline
     Low EGR Condition &   \makecell{3. \textit{accelpedpos}:  $a_{9}$ $a_{9}$ $a_{9}$ \\ \textit{EGRkgph}: $g_{0}$ $g_{0}$ $g_{0}$ \\} & \makecell{4. \textit{EGRkgph}: $g_{1}$ $g_{1}$ $g_{1}$ }\\
     \hline
     High Load Condition&   \makecell{5. \textit{EngTq}:  $L_{10}$ $L_{10}$ $L_{10}$ \\ \textit{engRPM} $N_{7}$ $N_{7}$ $N_{7}$ \\} & \makecell{6. \textit{accelpedalpos}:  $a_{8}$ $a_{8}$ $a_{8}$ \\ \textit{intakeP} $P_{8}$ $P_{8}$ $P_{8}$ }\\
      \hline
    \end{tabular}%

    }
         

      \label{tab:cooccurrence}
 \end{table}
 
\subsubsection{Experiment Design}
Experiments for this methodology were conducted using four sets of parameters as described in Table \ref{tab:parameterset}. Set1 contains all the seven attributes from the OBD dataset that were used in the physics-based prediction of NO\textsubscript{x} values as described in the previous section. Set2 contains the approximated time derivatives of these seven attributes over the dataset to get a better understanding of transient effects. Set3 contains 5 attributes that are not used in the physics model but are generally suspected to influence the emissions in a vehicle. Set4 contains the time derivatives of the five attributes used in the third set in order to understand the effect of transients in these values. 

\begin{table}[h!]
        \caption{Parameter Sets used in DWC Pattern Detection}
    \centering

    \begin{tabular}{c|m{9cm}}
    \hline
        \textbf{Parameter Set} & \textbf{Parameters Present} \\
        \hline
        Set 1 & \textit{engRPM, EGRkgph, AirInkgph, intakeT, intakeP, Fuelconskgph}\\
        Set 2 & \textit{engRPMdelta, EGRkgphdelta, AirInkgphdelta, intakeTdelta, intakePdelta, Fuelconskgphdelta}\\
        Set 3 & \textit{accelpedalpos,ExhaustT, Wheelspeed, EngTq, InstFuelCon}\\
        Set 4 & \textit{accelpedalposdelta,ExhaustTdelta, Wheelspeeddelta, EngTqdelta, InstFuelConsdelta}\\
        \hline
    \end{tabular}

    \label{tab:parameterset}
\end{table}

The output DWC patterns are fed into a post-processing step to identify the specific windows that are contained in each pattern and to sort them by their respective Cross-K function values. An analysis is next performed on whether a certain pattern contains predominantly under-predicted values (Predicted value less than observed value or data point below the 1-1 line in Figure \ref{fig:proppred}) or over-predicted values (opposite of under-prediction). This was done to gain insights into the magnitude-level performance of the proposed physics model. The Cross-K function lower bound threshold $\epsilon$ was set to 2.5, the $minsupp$ value to 0.01, the $summationThreshold$ to 30 ppm, and the window length $L$ to 3 seconds.

\subsubsection{Results of DWC Pattern Detection}
\begin{figure}[h!]
    \centering
    \includegraphics[width=0.5\linewidth]{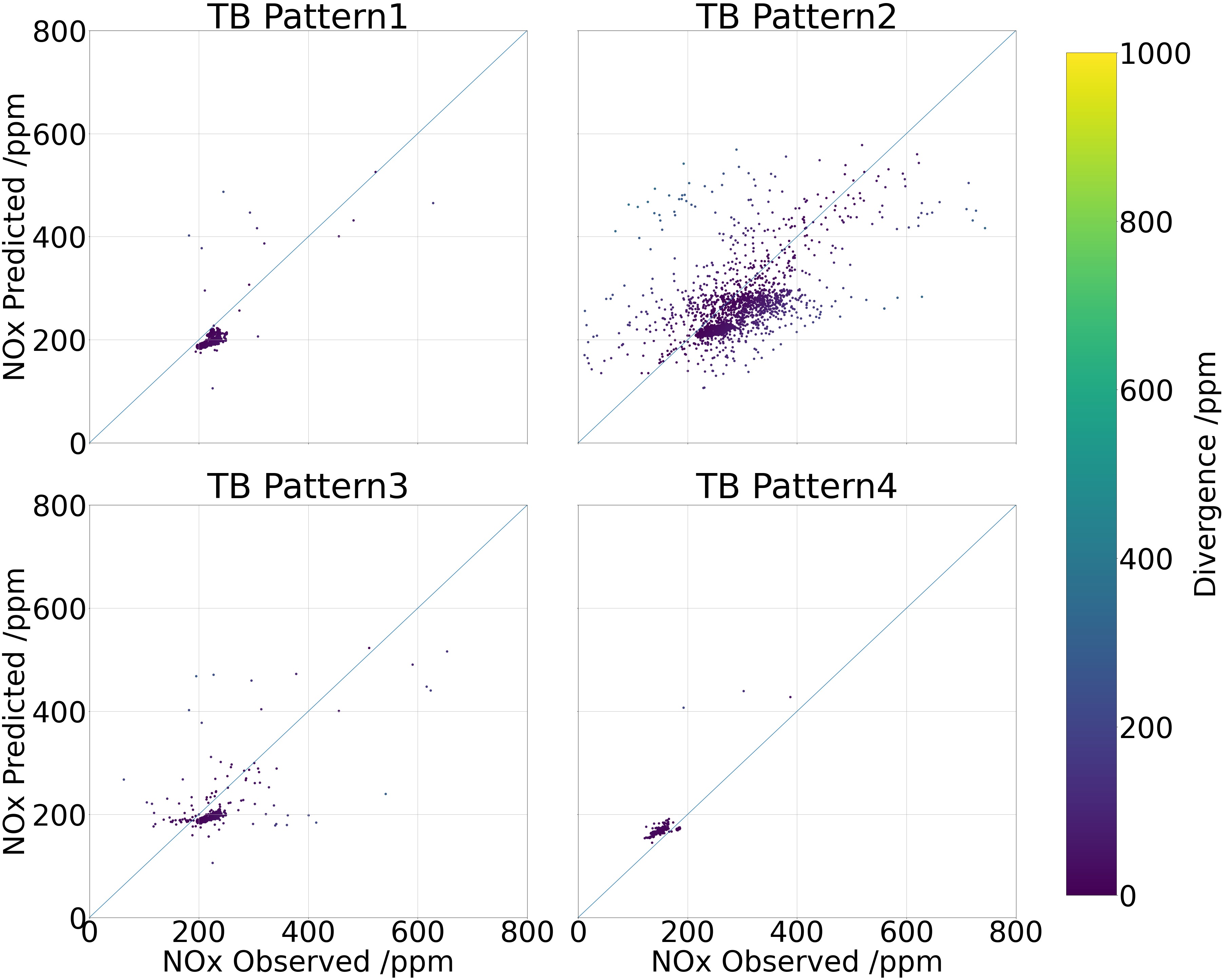}
    \caption{Scatter plots of the patterns identified in Table \texorpdfstring{\ref{tab:sigpatternsTB}}}
    \label{fig:sigpatternsTB}
\end{figure}

Figure \ref{fig:sigpatternsTB} shows the scatter plots of the four most statistically significant patterns that were identified using the DWC Pattern Detection Algorithm. A pattern with a higher Cross-K function value is considered to be more statistically significant. The parameters that co-occur in each of the patterns are provided in Table \ref{tab:sigpatternsTB}, along with the respective Cross-K function value, Count (Number of windows present in each pattern) and Support. The letters $A, T, P$, etc along with the subscripts are used to denote the magnitude of different attributes in a DWC pattern.  It can be observed that the significant patterns mentioned in the table belong to different representations of Engine Idling condition. The divergent windows that make up these chosen patterns are mutually exclusive. However, it is important to note that the divergent windows of some patterns identified by the algorithm might also be present in other patterns.  

\begin{table}[htp!]
    \caption{Description of the top four most statistically significant DWC patterns using parameters Set1 for the Transit Bus dataset}
    \label{tab:sigpatternsTB}
    \centering
    \begin{tabular}{c|l|r|r|r}
  	\hline

      \textbf{Pattern} & \textbf{Co-occurrence} & \textbf{Cross-K}& \textbf{Count}& \textbf{Support}\\
      \hline
       Pattern 1  & \makecell[l]{\textit{AirInkgph}: $A_1$ $A_1$ $A_1$\\ \textit{IntakeT}: $T_5$ $T_5$ $T_5$\\ \textit{IntakeP}: $P_0$ $P_0$ $P_0$ \\\textit{Fuelconskgph}: $F_1$ $F_1$ $F_1$}  & 3.0273 & 1453 & 0.0146 \\
       \hline
       Pattern 2 & \makecell[l]{\textit{EGRkgph}: $E_2$ $E_2$ $E_2$\\ \textit{Fuelconskgph}: $F_1$ $F_1$ $F_1$}& 2.9808 & 2954 & 0.0297 \\
       \hline
       Pattern 3 & \makecell[l]{\textit{EGRkgph}: $E_3$ $E_3$ $E_3$\\\textit{IntakeT}: $T_5$ $T_5$ $T_5$\\\textit{Fuelconskgph}: $F_1$ $F_1$ $F_1$} & 2.8678 & 1011 & 0.0102 \\ 
       \hline
      Pattern 4 & \makecell[l]{\textit{AirInkgph}: $A_0$ $A_0$ $A_0$\\ \textit{IntakeT}: $T_6$ $T_6$ $T_6$\\\textit{IntakeP}: $P_0$ $P_0$ $P_0$\\ \textit{Fuelconskgph}: $F_1$ $F_1$ $F_1$} & 2.7525 & 1227 & 0.0123\\  
      \hline
     \end{tabular}%
    \end{table}
    
\subsubsection{Complete Framework for \texorpdfstring{NO\textsubscript{x}}~ Emission Analysis}
\begin{figure}[ht]
    \centering
    \includegraphics[width=0.5\linewidth]{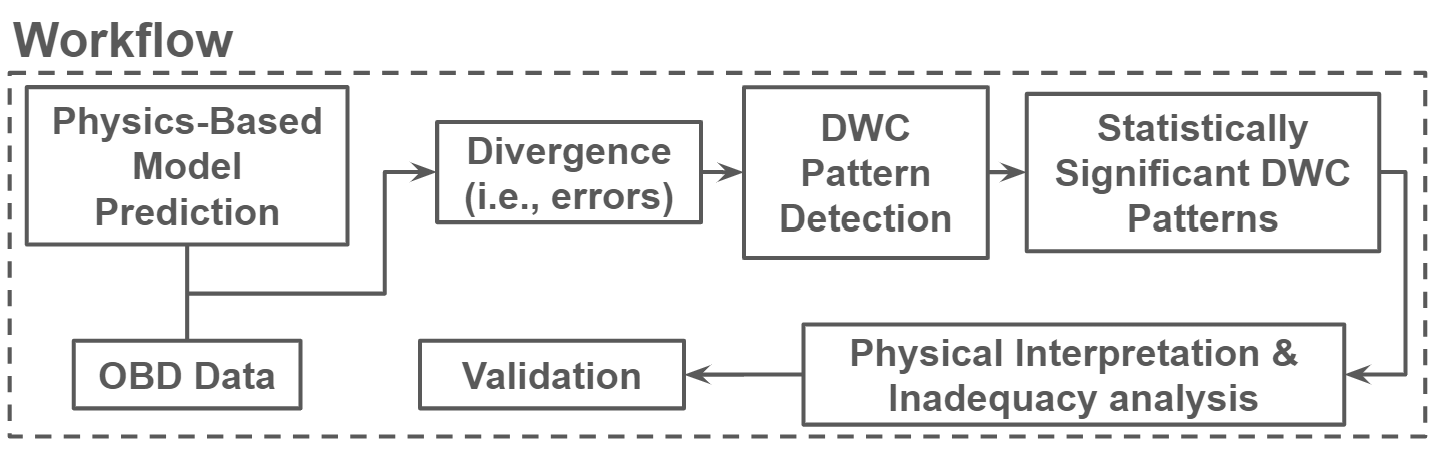}
    \caption{Complete framework for NO\textsubscript{x} value prediction}
    \label{fig:framework}
\end{figure}

The complete framework for predicting and analyzing NO\textsubscript{x} emission values from an OBD dataset is summarized in Figure \ref{fig:framework}. The OBD dataset is first used to estimate a time series of NO\textsubscript{x} emission values using a physics equation and a non-linear regression model. Divergences or errors of prediction are inputs into the DWC Pattern Detection Algorithm along with the attributes that are suspected to influence the formation of NO\textsubscript{x} in a CI engine. The algorithm outputs the statistically significant DWC patterns that are then processed to identify the constituent time windows in each pattern. An analysis of the four-parameter sets and a subsequent under- vs over- prediction algorithm provides an idea of how NO\textsubscript{x} is produced in a candidate engine in real-world conditions. The interesting results of the short-comings of the Physics-Based model in specific vehicle conditions are also examined.

\section{Validation of the Proposed Framework}
\subsection{Generalizability of the Framework}
To prove the generalizability of the proposed framework of NO\textsubscript{x} emissions prediction and analysis, an OBD dataset of another vehicle was anaylzed using the developed methodology. The data was obtained from \citet{FleetDNA}. The comparison vehicle chosen (due to availability of data) was a food delivery box truck fitted with a 2013 Peterbilt PX-7 (6.7L) engine operating in Denver, Colorado. The OBD dataset contained 101741 data entries of similar engine and vehicle attributes that were present in the transit bus dataset at 1Hz resolution. The data was measured over 206 trips taken by the truck over 2 months. The data was acquired using a commercially available OBD data logger (ISAAC Inc.). 

\begin{figure}[h!]
    \centering
    \includegraphics[width=0.5\linewidth]{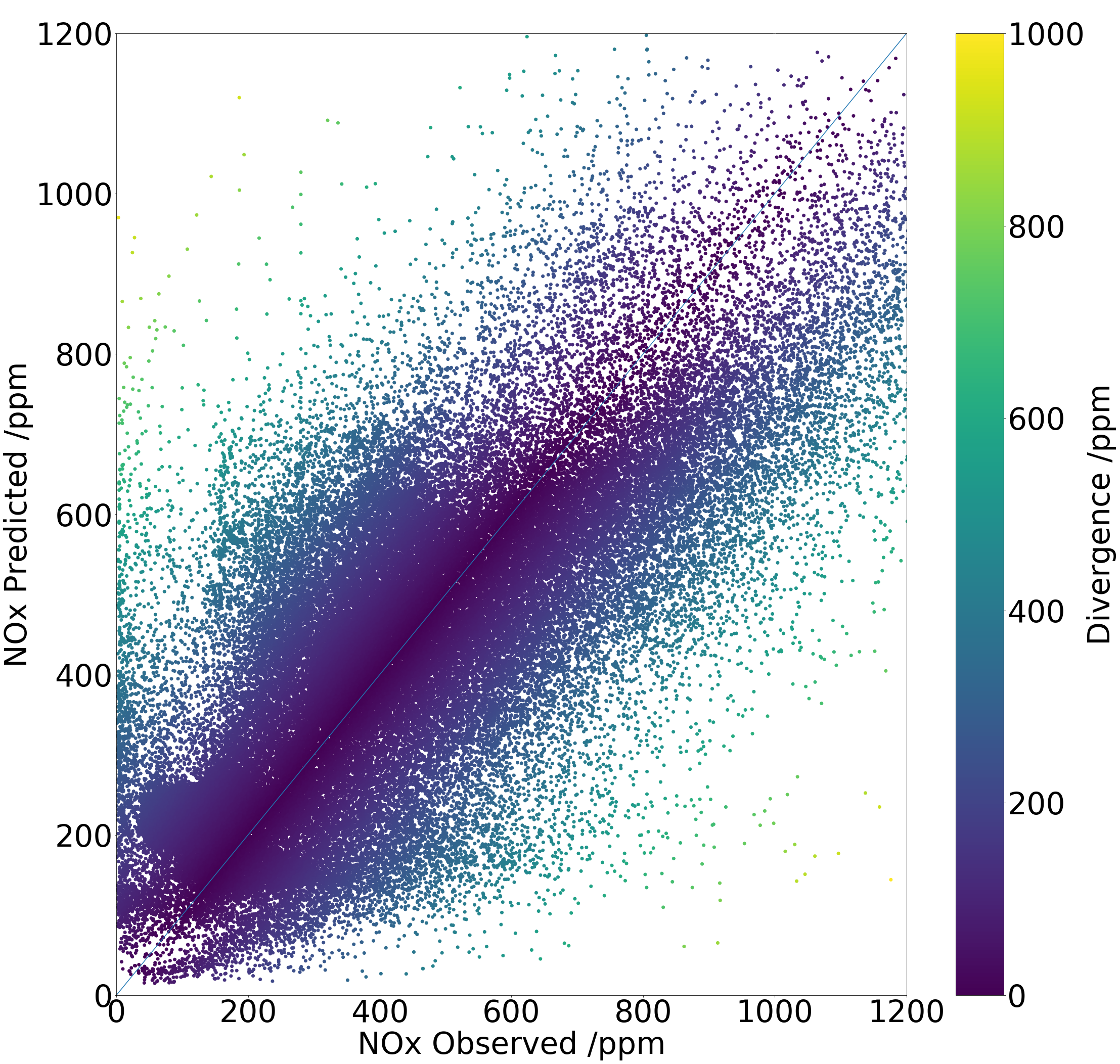}
    \caption{NO\textsubscript{x} emission prediction using the Proposed Physics-based Model for food delivery truck}
    \label{fig:predFDT}
\end{figure}

\begin{figure}[h!]
    \centering
    \includegraphics[width=0.5\linewidth]{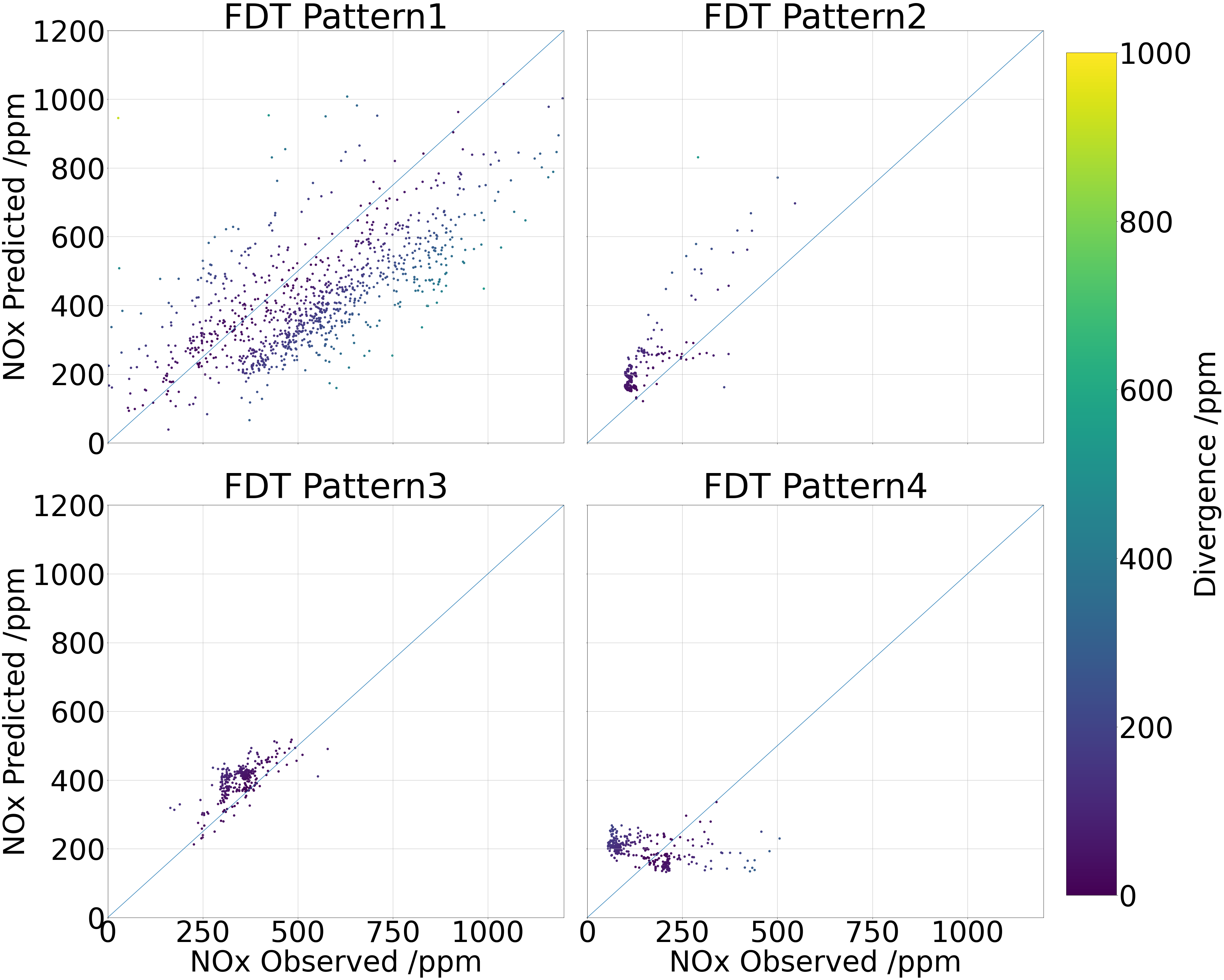}
    \caption{$NO_{x}$ predictions of the top four most statistically significant DWC patterns with parameters of Set1 for the food delivery truck dataset}
    \label{fig:sigpatternsFDT}
\end{figure}

Figure \ref{fig:predFDT} shows the scatterplot of NO\textsubscript{x} emission values for the food delivery truck predicted using the proposed physics model from Equation \ref{eq:propphysics}. Table \ref{tab:predaccCompare} compares the regression coefficients and the prediction accuracy of the emissions from the two vehicles. It can be observed that most of the regression coefficients are similar to the ones from the transit bus dataset which are in turn very close to the expected values from the chemical rate equation for the Extended Zel'dovich Mechanism. Figure \ref{fig:sigpatternsFDT} shows the four most significant patterns that were mined from the food delivery truck dataset. The significant DWC patterns are quite different because the operating conditions of the food delivery truck are quite different from that of the transit bus. A transit bus is generally operated between nearby bus stations with varying payloads due to passengers boarding and leaving the vehicle. A food delivery truck carries a mainly consistent payload and makes longer trips on highways. The similar magnitudes of the regression coefficients and the predictive accuracy for the two datasets validates the generalizability of the proposed framework and shows that it could also be applicable to other compression-ignition engine powered vehicles. 

\begin{table}[h!]
    \caption{Comparison of predictive accuracy of the proposed physics-based model(PBM)\nomenclature[A]{PBM}{Physics-based Model} and deep neural network(DNN) models for the transit bus(TB) \nomenclature[A]{TB}{Transit bus} and food delivery truck(FDT)\nomenclature[A]{FDT}{Food delivery truck} datasets}
    \centering
    \begin{tabular}[scale=\linewidth]{l|c|c|c|c}
    \hline
        \textbf{Model} & \multicolumn{1}{|c|}{\textbf{R\textsuperscript{2}}} & \textbf{RMSE} & \textbf{MAE}& \textbf{Coefficients [a, b, c, d, e]}\\
         & & \textbf{(ppm)} & \textbf{(ppm)} & \\
        \hline
        PBM-TB & 0.6175 & 92.37 & 68.70 & 2.15e-3, 0.013, 0.36, 0.082, 2117.33 \\
        PBM-FDT & 0.6004 & 168.45 & 124.06 & 7.46e-4, 0.235, 0.74, 0.820, 732.51 \\
        DNN-TB & 0.7103 & 81.05 & 54.67 & -\\
        DNN-FDT & 0.7214 & 152.04 & 100.40 & -\\
        \hline
    \end{tabular}
    \label{tab:predaccCompare}
\end{table}

\subsection{Sensitivity Analysis of the Proposed Physics-Based Model}
\begin{figure}[h!]
        
        \centering
        \includegraphics[width=0.5\linewidth]{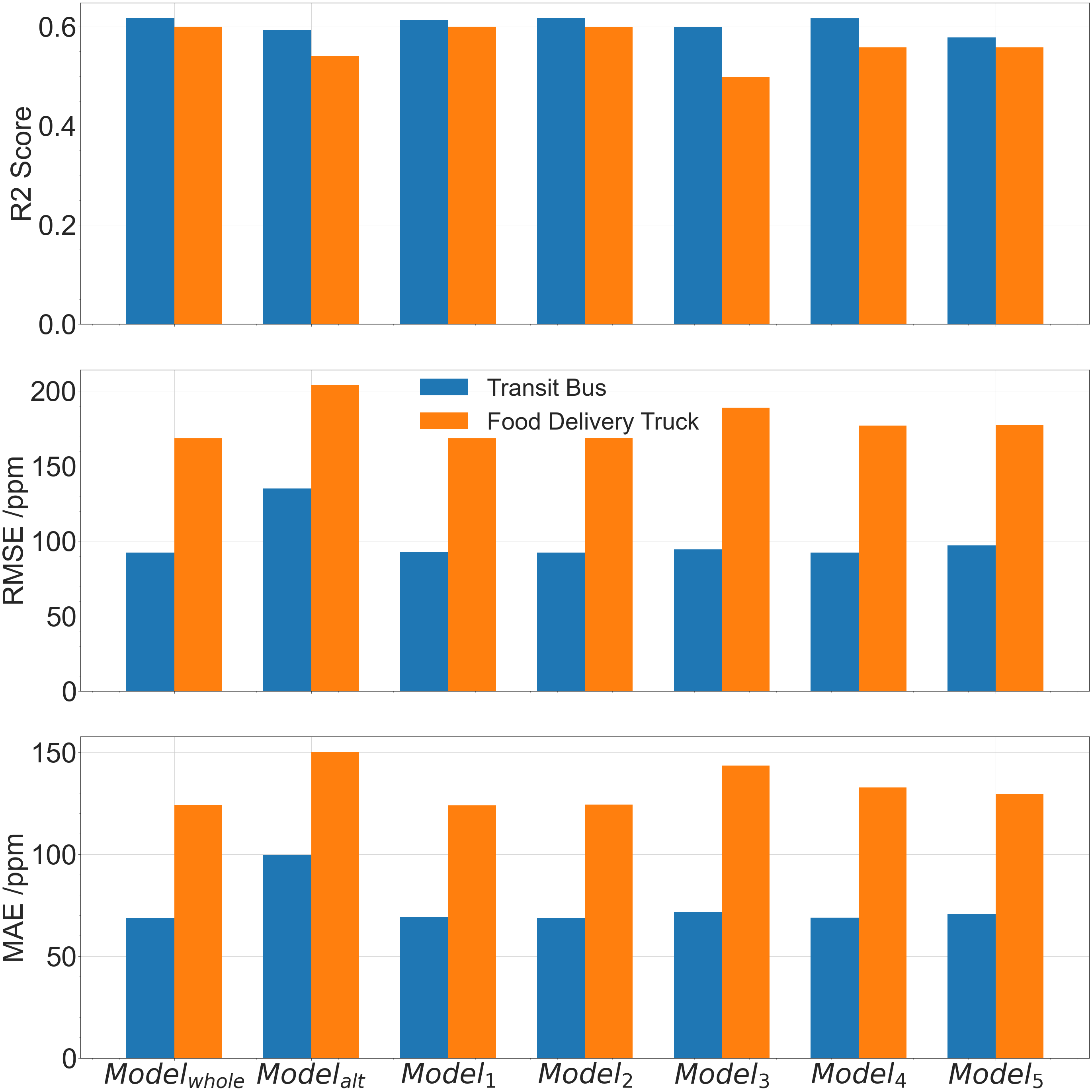}
      \caption{Bar charts of predictive accuracy for the different models given in Table \ref{tab:sensphysics} that were evaluated on the transit bus dataset}
    \label{fig:sensphysics}
    \end{figure}

The proposed physics-based model equates $x_{NO_x}$ with three terms($t_{res}$, $x_{O_2}$ and $T_{adiab}$) calculated from the engine parameters present in the OBD dataset. The reason for choosing this model is that it is the most similar to the general rate equation for the formation of NO\textsubscript{x} using the Extended Zel'dovich Mechanism. An analysis was performed to find the extent of dependence of $x_{NO_x}$ on the three terms by performing non-linear regression with two terms at a time for the both the TB and FDT dataset. Table \ref{tab:sensphysics} shows the equations for the reduced models in consideration for this analysis on both the datasets. The different models were obtained through a drop-column method. This includes the original model as described in Equation \ref{eq:propphysics} which is called \textit{Model\textsubscript{whole}}. \textit{Model\textsubscript{alt}} is the original model but the regression coefficients (a,b,c,d, and e) are substituted with the values obtained from the proposed physics-based model for the other dataset. That is, the \textit{Model\textsubscript{alt}} for the Transit Bus data uses coefficients obtained from the food delivery truck dataset, and vice versa. The variation in the predictive accuracy metrics are plotted in Figure \ref{fig:sensphysics}. 

As shown in Figure \ref{fig:sensphysics} the two datasets show a similar trend for most of the models under consideration. \textit{Model\textsubscript{alt}} gives less desirable predictive accuracy for both the datasets, leading to the conclusion that the engine characteristics and operations are different for the two vehicles. Each vehicle data requires its own set of regression coefficients to provide a good predictive accuracy. \textit{Model\textsubscript{5}}, the only model that does not include $T_{adiab}$, gives the worst predictive accuracy for both the datasets (except for the $R^2$ value for FDT). This suggests that $T_{adiab}$ is the most important term (among the terms considered) that determines NO\textsubscript{x} formation in a compression-ignition engine. \textit{Model\textsubscript{4}}, which is the only model that doesn't consider $t_{res}$, gives the best predictive accuracy apart from \textit{Model\textsubscript{whole}}. This suggests that  $t_{res}$ affects NO\textsubscript{x} formation the least significant amount among the three terms used in the physics equation. 
\begin{table}[h!]
        \caption{Different candidate equations obtained via drop-column method on Equation \texorpdfstring{\ref{eq:kinetic}}}
        \centering
    \begin{tabular}{c|m{10cm}}
    \hline
        \textbf{x\textsubscript{NO\textsubscript{x}} Model} & \textbf{Equation} \\\hline
        \textit{Model\textsubscript{whole}} & $a*(\hat{t}_{comb})^b*x_{O_2}^c*\hat{T}_{adiab}^d*exp(-e/T_{adiab})$\\
        \textit{Model\textsubscript{alt}} & $a*(\hat{t}_{comb})^b*x_{O_2}^c*\hat{T}_{adiab}^d*exp(-e/T_{adiab})$ using coefficients from the FDT dataset\\
        \textit{Model\textsubscript{1}} & $a*(\hat{t}_{comb})^b*x_{O_2}^c*\hat{T}_{adiab}^d$\\
        \textit{Model\textsubscript{2}} & $a*(\hat{t}_{comb})^b*x_{O_2}^c*exp(-e/T_{adiab})$\\
        \textit{Model\textsubscript{3}} & $a*(\hat{t}_{comb})^b*\hat{T}_{adiab}^d*exp(-e/T_{adiab})$\\
        \textit{Model\textsubscript{4}} & $a*x_{O_2}^c*\hat{T}_{adiab}^d*exp(-e/T_{adiab})$\\
        \textit{Model\textsubscript{5}} & $a*(\hat{t}_{comb})^b*x_{O_2}^c$\\
        \hline
    \end{tabular}
    \label{tab:sensphysics}
\end{table}

\subsection{Comparison with Deep Neural Network Model}
\begin{figure*}[htbp]
     \centering
     \begin{subfigure}[b]{0.45\linewidth}
         \includegraphics[width=\linewidth]{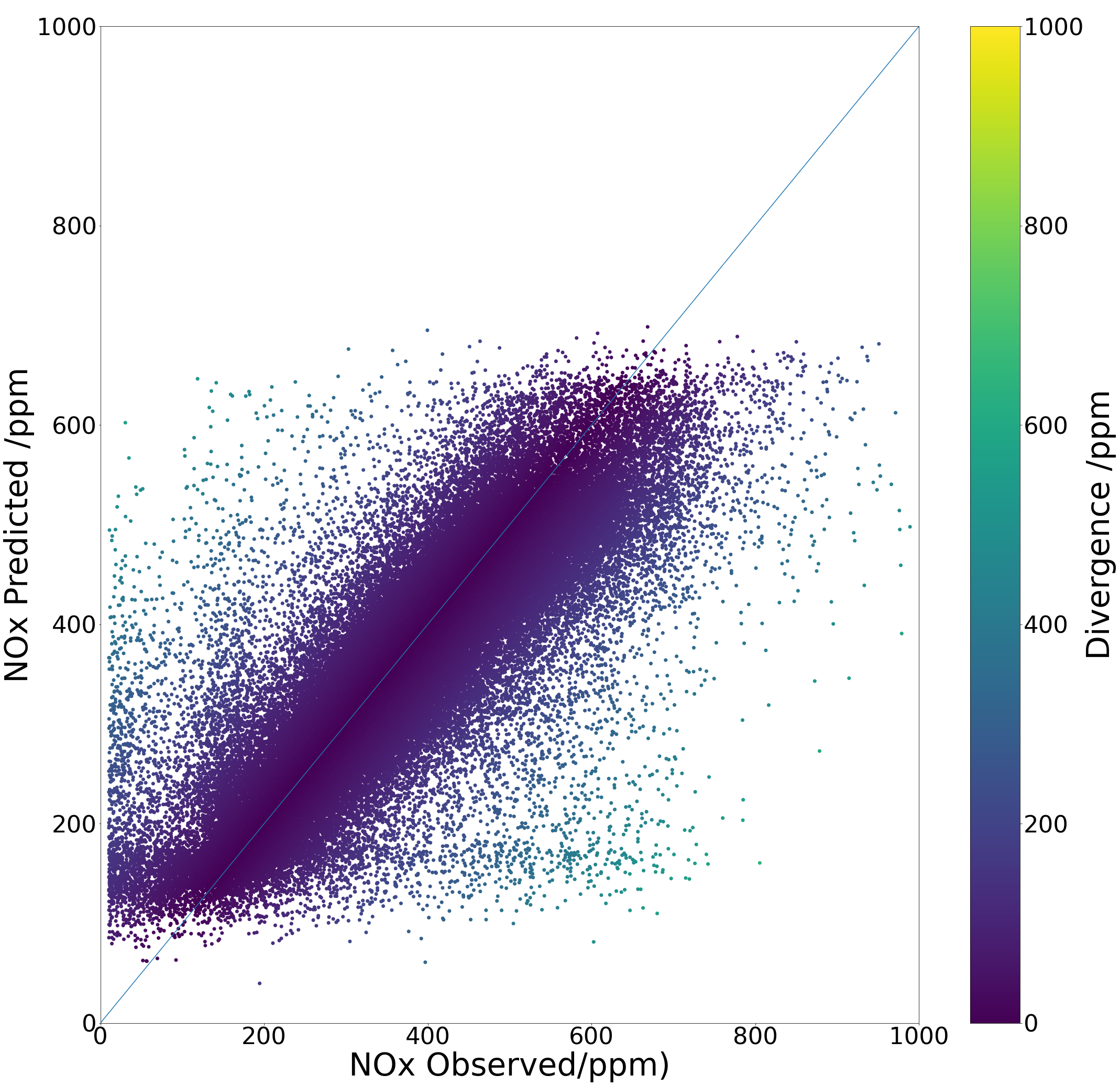}
         \caption{Transit Bus}
     \end{subfigure}
    ~
     \begin{subfigure}[b]{0.45\linewidth}
         \includegraphics[width=\linewidth]{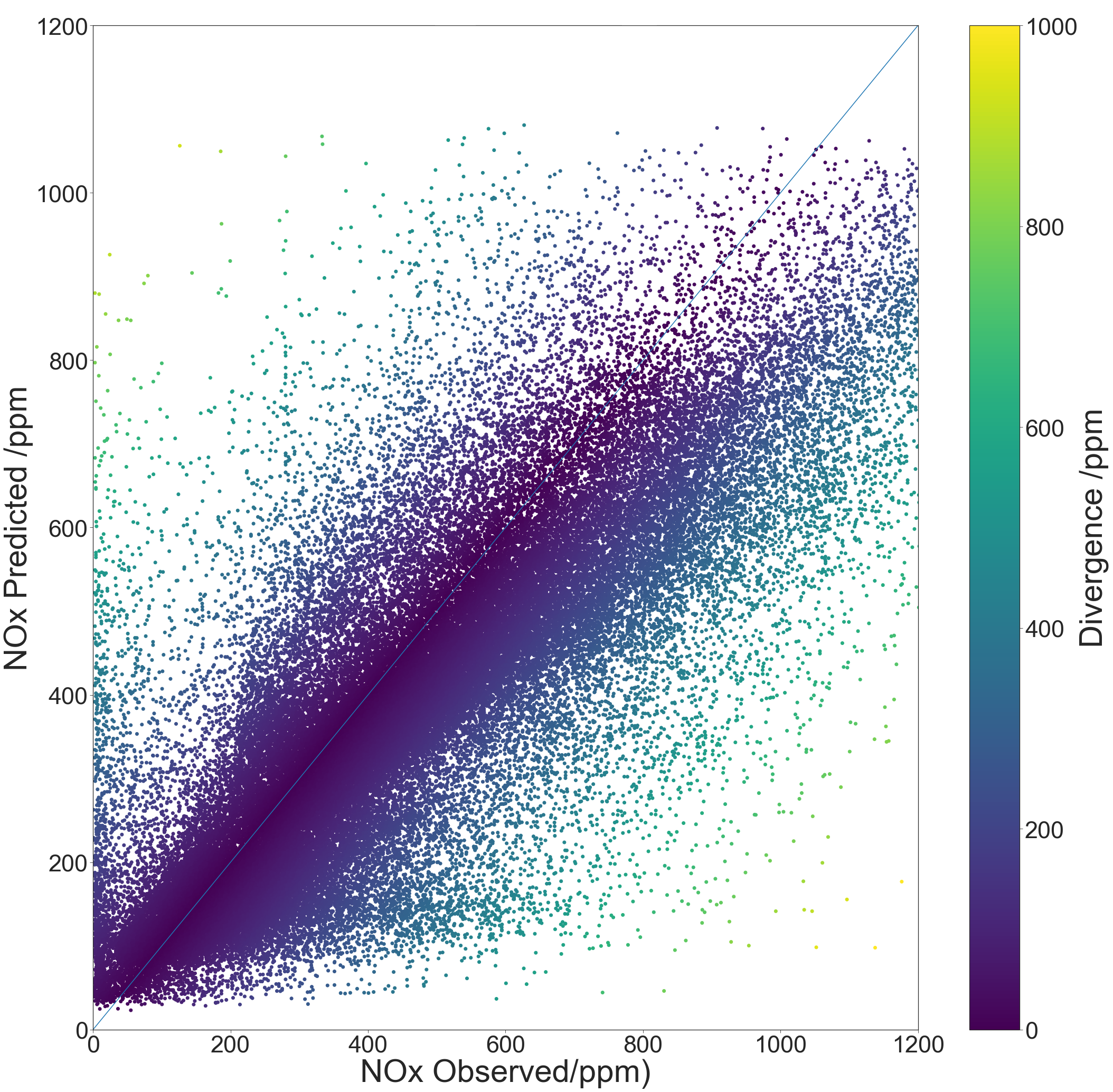}
         \caption{Food delivery truck}
     \end{subfigure}
     \caption{Deep Neural Network prediction of NO\textsubscript{x} emissions for the two datasets}
      \label{fig:DNNpred}
    \end{figure*}

A deep neural network(DNN)\nomenclature[A]{DNN}{Deep neural network} is a black box model that contains multiple hidden layers along with input and output layers. For prediction of NO\textsubscript{x} emission values, a multi-layer perceptron model (MLPRegressor) from the Scikit-Learn Python package \citet{scikit-learn} was employed. The input nodes contain the three terms used in the proposed physics model ($t_{res}$, $x_{O_2}$, and $T_{adiab}$) and the output layer is a single node that regresses $x_{NO_x}$ values. Each dataset is split into the training and testing set in the ratio of 8:2 . The model uses 100 hidden layers with the ReLU activation function since this number of layers provide the best result without overfitting.  The scatterplot of the DNN based prediction for both the vehicle datasets are shown in Figure \ref{fig:DNNpred}. The predictive accuracy using DNNs are listed in Table \ref{tab:predaccCompare}. As expected, all three metrics are significantly better compared to the proposed physics-based model prediction.

The advantage of the DNN method is its high predictive accuracy. As a black-box model, the major disadvantage of the DNN is that it provides little information about how the three terms are dependent in order to output $x_{NO_x}$ values. The approach trades some accuracy for interpretability and transparency (\citet{BARREDOARRIETA202082}). The proposed framework consisting of the physics-based model and the DWC Pattern Detection algorithm is transparent because the regression coefficients can be obtained from the model. The framework is interpretable because how to recreate the predictions manually if required can be understood, and it is explainable since the regression coefficients can be compared to empirical correlations found in the literature. The DNN method has high predictive accuracy, but like all black-box models, it does not inform the user how the output emissions values are obtained. Thus, which relationships among the three terms are most effective at describing NO\textsubscript{x} formation in an engine cannot be understood. 

\section{Discussion and Conclusions}
The proposed framework for predicting NO\textsubscript{x} emission values from an OBD dataset using divergent window co-occurrence pattern detection to detect data patterns that correspond to scenarios in which the general model might be inadequate is a good way to understand the formation of NO\textsubscript{x} emissions inside a compression-ignition engine operating in real-world conditions. Although low cost sensors contain less accurate data when compared to laboratory grade sensors used in engine dynamometer experiments, results show that vehicle on-board diagnostics datasets are extremely useful in understanding emissions trends during real-world operating conditions of the vehicle. The proposed physics model also has better predictive accuracy compared to the baseline and regressed baseline models. 

Regression coefficients obtained from the proposed physics model used to predict NO\textsubscript{x} emission values are similar to those obtained through laboratory experiments on the Extended Zel'dovich mechanism of formation of nitrogen oxides. This justifies the use of the approximated parameters {$\hat{t_{res}}$, $\hat{T_{adiab}}$, } and {$x_{O_2}$} in the prediction of NO\textsubscript{x}.   The varying magnitudes of divergence of predicted values from the observed values confirms that a one-size-fits-all model is not sufficient to express the different operating conditions of a vehicle in real operation. The formation of NO\textsubscript{x} in a diesel vehicle is affected by many factors such as driver controls, environmental conditions, engine control systems and so on. Inadequacies of the general physics-based model can be identified by clustering these varied operating conditions into patterns using the proposed DWC Pattern Detection algorithm. Any time windows that are not identified through the algorithm can be assumed to be adequately explained by the proposed general physics model. An understanding of when and how such vehicle operating conditions occur is important in applications such as identifying emissions defeat devices, testing emissions regulations, and also diagnosis of engine components. An example of a diagnostic analysis is if the NO\textsubscript{x} emissions predicted using the proposed physics-based model for a specific vehicle does not correlate well with the certain feature variables that define the functioning of a component, then the specific component can be identified to be faulty and may require replacement.

An important observation made while analyzing the dataset using the DWC Pattern Detection Algorithm is that transient patterns are not statistically significant. A transient pattern is one where an attribute value is not constant throughout a candidate time window. The absence of transient patterns might be due to two reasons: the dataset contains too many different permutations of attributes values in time windows of the chosen window length, or NO\textsubscript{x} predicted during transients are well described by the general physics-based model and has low divergence.

This paper proposes a novel framework for predicting and analyzing NO\textsubscript{x} emissions values using a physics-based model and the Divergent Window Co-occurrence(DWC) Pattern Detection algorithm, given a vehicle's on-board diagnostics dataset. The results show that the proposed physics model predictions give around $55\%$ better root mean square error, and around $60\%$ mean absolute error compared to the baseline NO\textsubscript{x} prediction model. The Divergent Window Co-occurrence Pattern Detection algorithm identified clusters of time windows in the OBD dataset that have large prediction errors or divergence. The algorithm identified statistically significant patterns such as engine idling conditions where the physics model is inadequate in describing NO\textsubscript{x} formed during a specific vehicle operating condition. The framework was first used on an OBD dataset of a transit bus and the generalizability of the model was validated on a dataset from a food delivery truck. The dependence of the three terms in the proposed physics model was examined. A sensitivity analysis of the three terms present in the physics-based model indicated that $T_{adiab}$ is the most important factor in describing NO\textsubscript{x} formation in an engine, and the least important factor among the three is $t_{res}$. A comparison of the physics-based model with a deep neural network shows that the latter has a higher predictive accuracy but lacks in interpretable and transparency. The results from the prediction are observed to behave in a similar manner for both datasets and can be used to prove that NO\textsubscript{x} formation happens in a similar manner in different powertrain setups. The varying results from the pattern detection analysis for the two datasets prove that the variations in emissions are a function of how the vehicle is operated in real-world conditions. 

\section{Data Availability}
The on-board diagnostics (OBD) data used in this paper can be accessed at the Github Repository at \href{https://github.com/bharatwrrr/engine-nox-data.git}{https://github.com/bharatwrrr/engine-nox-data.git}

\section{Funding}
This material is based upon work supported by the National Science Foundation under Grant No. 1901099. 
\section{Acknowledgements}
  The authors thank the U.S. Department of Energy’s National Renewable Energy Laboratory for their FleetDNA support and assistance. The views and opinions of authors expressed herein do not necessarily state or reflect those of the United States Government or any agency thereof.



\newpage

\appendix
\section{Spatio-temporal terminology: Ripley's Cross-K Function}

\begin{figure*}[htbp!]
     \centering
     \begin{subfigure}[b]{0.3\linewidth}
         \includegraphics[width=\linewidth]{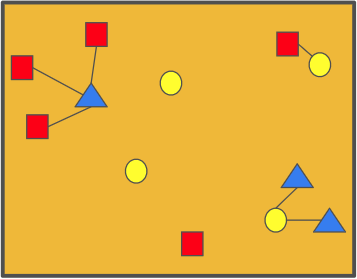}
         \caption{Spatial data points with connected neighbors}
         \label{fig:appendix_eg1}
     \end{subfigure}
     \begin{subfigure}[b]{0.3\linewidth}
         \includegraphics[width=\linewidth]{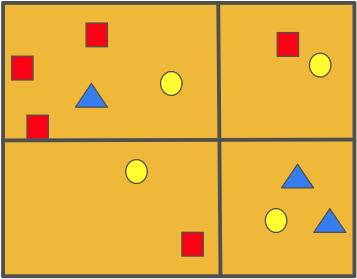}
         \caption{Spatial data points in a partitioned data space}
         \label{fig:appendix_eg2}
     \end{subfigure}
     \begin{subfigure}[b]{0.3\linewidth}
         \includegraphics[width=\linewidth]{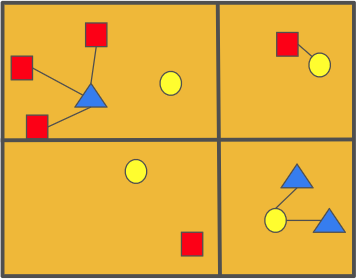}
         \caption{Spatial data points indicating neighbors and partitioned data space}
         \label{fig:appendix_eg3}
     \end{subfigure}
     \caption{An example of a spatial data set.}
      \label{fig:appendix_eg}
    \end{figure*}

Ripley’s K function is a spatial analysis tool used to describe how point patterns occur over a given area of interest. Ripley’s K allows researchers to determine if the phenomenon of interest appears to be dispersed, clustered, or randomly distributed throughout the study area. 
\newline
Using an illustrative example, the Ripley’s K function is used to describe the point pattern of three different kinds of spatial objects, represented by squares, triangles and circles in Figure \ref{fig:appendix_eg1}. The function accounts for the number of object pairs that can be deemed neighbors. In a spatial context, the classification of an object pair as neighbors is typically done by a Boolean Distance Buffer (\citet{koperski_discovery_1995}) using which objects are deemed neighbors if the distance between them is less than a threshold. The number of observed neighboring pairs is then traditionally compared to the number of neighboring pairs one would expect to find based on a completely spatially random point pattern. This example analyses the object pair square (red) - triangle (blue). Fulfilling an arbitrarily chosen distance-based criterion for being neighbors, the data set displays 3 neighboring pairs of a square and a triangle. In a completely spatially random distribution, one would expect to locate a total of 15 neighboring pairs of a square and a triangle. Hence, the value of Ripley's cross-K function is the ratio of the two values, i.e., $0.2$.\par
Theoretically the K-function is defined as,
\begin{equation}
K(h, data)=\lambda\textsuperscript{-1}E[D]
\label{eq:K-func}
\end{equation}
where $\lambda$ is the density of events (number of events per unit area) and D is the number of  events within distance h of an arbitrary event \citet{ripley_second-order_1976}.

In this research work, the form of the Cross-K function used is a purely temporal one as described by Ali et. al. \cite{Ali2017}. It is used to measure the association between a non-compliant window co-occurrence pattern of a combination of input variables and the occurrence of non-compliant windows of the target variable, at a specified time lag. Equation \ref{eq:K-func} is then transformed into,
\begin{equation}
K\textsubscript{C,W\textsubscript{N}}(\delta)=\lambda\textsubscript{W\textsubscript{N}}\textsuperscript{-1}E[P]
\label{eq:K-func-temporal}
\end{equation}
where $K\textsubscript{C,W\textsubscript{N}}(\delta)$ is the value of the K function measuring the association between a non-compliant window co-occurrence pattern, C, and the occurrence of  non-compliant windows at a time lag $\delta$, P is the number of number of non-compliant windows $W\textsubscript{N}$ starting within time $\delta$ from the start of an instance of C, and $\lambda\textsubscript{W\textsubscript{N}}$ is the expected number of non-compliant windows per unit time. In the case of temporal randomness (an analogy of complete spatial randomness in the time domain), the value of $K\textsubscript{C,W\textsubscript{N}}(\delta)$ is equal to $(\delta+1)$, and values greater than $(\delta + 1)$ indicate strong association. 


\section{Spatio-temporal terminology: Support}
Support gives insights on the distribution of objects in a partitioned data space. Considered the partitioned space with 3 different kinds of spatial objects in Figure \ref{fig:appendix_eg2}. Let $P$ be the set of all possible predicates. Consider a rule $p\textsubscript{1}\xrightarrow{}p\textsubscript{2}$, where $p\textsubscript{1}$, $p\textsubscript{2}\in P$. An example of $p\textsubscript{1}$ can be a red object (square) and an example of $p\textsubscript{2}$ can be a blue object (triangle). Support for the rule $p\textsubscript{1}\xrightarrow{}p\textsubscript{2}$ is defined as the ratio of the number of objects (partitions) in the database that contain both $p\textsubscript{1}$ and $p\textsubscript{2}$ to the total number of objects (partitions) in the database. For the mentioned example (considering the red object and the blue object), only one out of the four partitions in the database contain both a red object and a blue object. Hence, the support for this example is $0.25$.

\newpage

\bibliographystyle{elsarticle-num-names} 
\bibliography{References.bib}



\end{document}